\documentclass{article}

\PassOptionsToPackage{round}{natbib}
\usepackage[final]{neurips_2022}

\usepackage[utf8]{inputenc} 
\usepackage[T1]{fontenc}    
\usepackage{hyperref}       
\usepackage{url}            
\usepackage{booktabs}       
\usepackage{amsfonts}       
\usepackage{nicefrac}       
\usepackage{microtype}      
\usepackage{xcolor}         
\usepackage{algorithm}
\usepackage[noend]{algpseudocode}
\usepackage{graphicx}
\usepackage{pgfplots}
\usepackage{tikz}
\usepackage[eulergreek]{sansmath}
\usepackage{setspace}
\usepackage{float}
\usepackage{subcaption}
\usepackage{amsthm}
\usepackage{multicol}
\usepackage{multirow}
\usepackage{wrapfig}
\usepackage{enumitem}

\usepackage{soul} 

\usepgfplotslibrary{fillbetween}
\pgfplotsset{
    compat=1.17,
    every tick label/.append style={font=\tiny\sansmath\sffamily},
    every axis label/.append style={font=\tiny\sansmath\sffamily},
    every axis plot/.append style={thick},
    legend style={font=\sansmath\sffamily},
}

\newtheorem{proposition}{Proposition}

\usepackage{color, colortbl}

\definecolor{napiergreen}{rgb}{0.16, 0.5, 0.0}

\usepackage{amsmath,amsfonts,bm}

\def\Apxref#1{Appendix~\ref{#1}}

\def\eqref#1{(\ref{#1})}

\def\1{\bm{1}}

\def\rvu{{\boldsymbol{i}}}

\def\rvs{{\boldsymbol{s}}}

\def\rvu{{\boldsymbol{u}}}
\def\rvv{{\boldsymbol{v}}}

\def\rvx{{\boldsymbol{x}}}
\def\rvy{{\boldsymbol{y}}}
\def\rvz{{\boldsymbol{z}}}

\def\mI{{\bm{I}}}

\def\mR{{\bm{R}}}

\DeclareMathAlphabet{\mathsfit}{\encodingdefault}{\sfdefault}{m}{sl}
\SetMathAlphabet{\mathsfit}{bold}{\encodingdefault}{\sfdefault}{bx}{n}

\newcommand{\E}{\mathbb{E}}

\newcommand{\R}{\mathbb{R}}

\newcommand{\KL}{D_{\mathrm{KL}}}

\title{A Rotated Hyperbolic Wrapped Normal Distribution for Hierarchical Representation Learning}

\author{%
  $\textrm{Seunghyuk Cho}^1$
  \And
  $\textrm{Juyong Lee}^1$
  \And
  $\textrm{Jaesik Park}^{1,2}$
  \And
  $\textrm{Dongwoo Kim}^{1,2}$
  \AND
  $\textrm{CSED POSTECH}^1$
  \And
  $\textrm{GSAI POSTECH}^2$
}

\begin{document}

\maketitle

\begin{abstract}
We present a rotated hyperbolic wrapped normal distribution (RoWN), a simple yet effective alteration of a hyperbolic wrapped normal distribution (HWN). The HWN expands the domain of probabilistic modeling from Euclidean to hyperbolic space, where a tree can be embedded with arbitrary low distortion in theory. In this work, we analyze the geometric properties of the \emph{diagonal} HWN, a standard choice of distribution in probabilistic modeling. The analysis shows that the distribution is inappropriate to represent the data points at the same hierarchy level through their angular distance with the same norm in the Poincar\'e disk model. We then empirically verify the presence of limitations of HWN, and show how RoWN, the proposed distribution, can alleviate the limitations on various hierarchical datasets, including noisy synthetic binary tree, WordNet, and Atari 2600 Breakout. The code is available at \url{https://github.com/ml-postech/RoWN}.

\end{abstract}
\section{Introduction}

Hyperbolic space has served as an effective medium to learn parsimonious representations of hierarchical data, including vocabulary with relationships~\citep{nickel17, nickel18, tifrea19}, knowledge graphs~\citep{chami20, sun20}, and social networks~\citep{zhao11, shavitt08}. Recent studies reveal that the underlying anatomy in much complex data is non-Euclidean, supporting the success of representation learning in hyperbolic space~\citep{bronstein17}.
However, due to the absence of well-defined distribution that is easy to sample and has an analytic density function in hyperbolic space, earlier studies have focused on the non-probabilistic learning framework.

Hyperbolic wrapped normal distribution (HWN) has been recently proposed as an alternative to the celebrated normal distribution in Euclidean space. With an analytic density function and easiness of sampling, the HWN expands the domain of probabilistic modeling from Euclidean to hyperbolic space.
The HWN has been successfully applied in various probabilistic models, including VAE~\citep{kingma14} and probabilistic word embeddings~\citep{vilnis15}. However, unlike the normal distribution in Euclidean space, the geometric characteristics of the HWN have not been fully understood so far. Therefore, figuring out what can be done or not with the HWN is difficult.

In this work, we analyze the geometric properties of the \emph{diagonal} HWN, a standard choice of distribution in many probabilistic models~\citep{mathieu19, nagano19}. Based on the observation that the principal axes of the diagonal normal distribution in Euclidean space have a parallel structure with the standard bases, we also focus on the structure of the principal axes of HWN in hyperbolic space to delve into a deeper understanding. Our analysis shows that the principal axes of the diagonal HWN are locally parallel to the standard bases in the Poincar\'e disk model. 

\begin{figure}[t!]
    \centering
    \begin{subfigure}[t]{.45\linewidth}
        \centering
        \includegraphics[width=.6\linewidth]{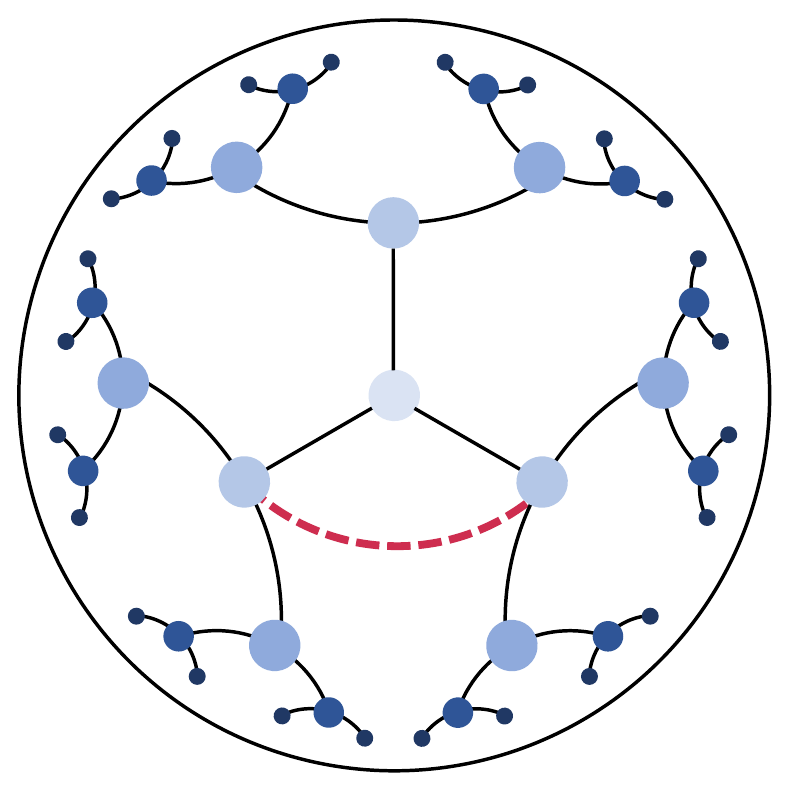}
        \caption{The Poincar\'e disk model}
    \end{subfigure}
    \begin{subfigure}[t]{.45\linewidth}
        \centering
        \includegraphics[width=\linewidth]{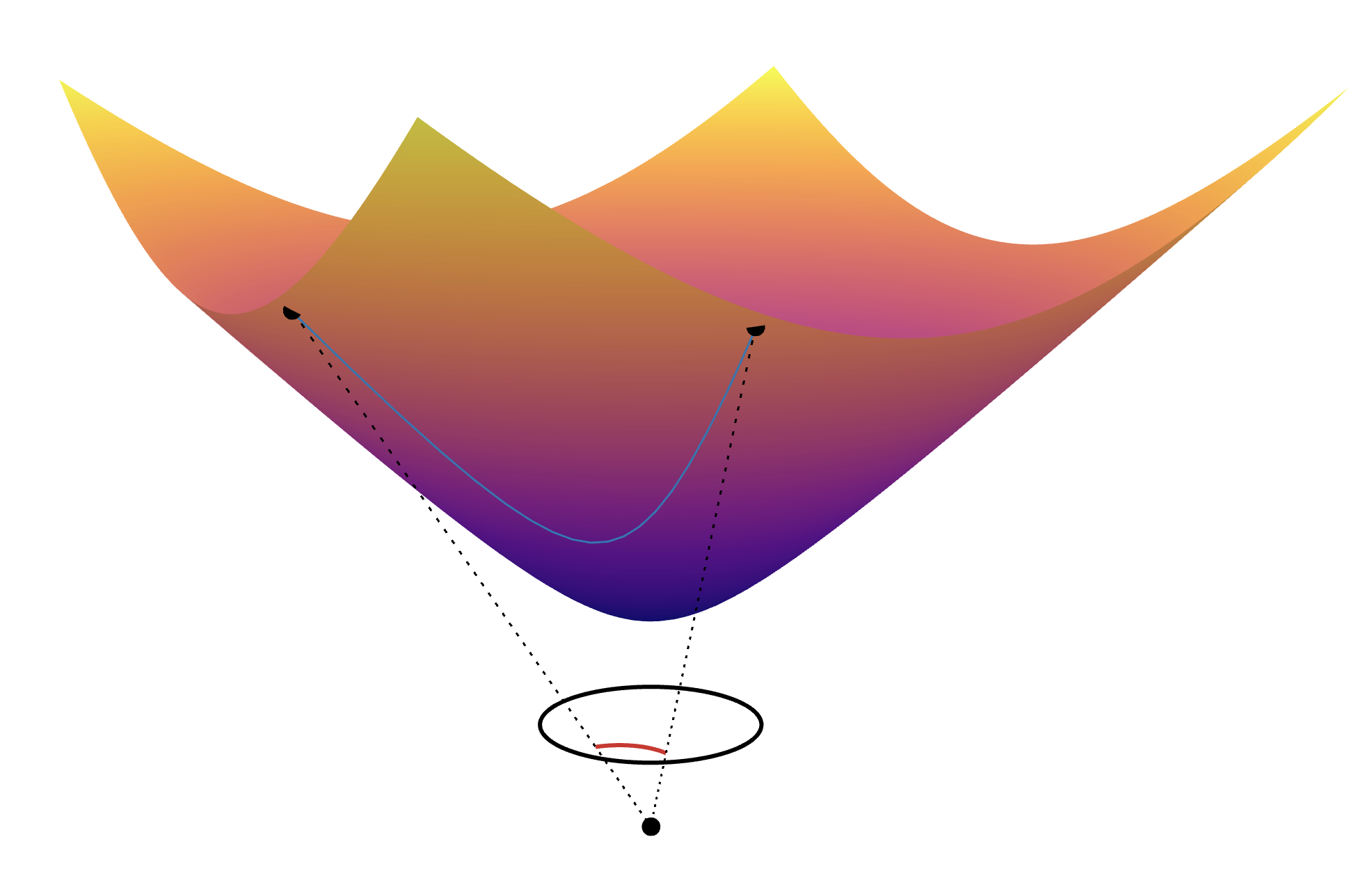}
        \caption{The Lorentz model}
    \end{subfigure}
    \hfill
    \caption{Visualization of hyperbolic space models. 
    (a) The Poincar\'e disk model can embed a given tree-structured data with low distortions as shown in the illustration. 
    The black segments refer to the shortest path between the points in the Poincar\'e disk model.
    The red dashed line denotes the continuous points in the same hierarchy level.
    (b) The Lorentz model is another model for hyperbolic space.
    The blue line is the shortest path between the points on the Lorentz model. 
    The red line is the projected geodesic via the diffeomorphism, which is still a geodesic of the Poincar\'e disk model.
    }
    \label{fig:hyperbolic_space}
\end{figure}

\autoref{fig:hyperbolic_space} (a) shows a common understanding on learned representation of hierarchical data in the Poincar\'e disk. The hierarchical structure spreads out like the spokes of a wheel. The local variation in a hierarchy can then be represented as an angular difference between nodes at the same level of the norm, i.e., the principal axis representing local variation is orthogonal to the radial axis. However, our analysis of the geometric property reveals that the local variation can only be modeled along with the standard bases with the diagonal HWN.

To fix the structure of the principal axes in the diagonal HWN, we propose a simple yet effective alteration of HWN, named a rotated hyperbolic wrapped normal distribution (RoWN). By rotating the diagonal covariance matrix before parallel transportation of HWN, we could resolve the limitations in local variation structure while keeping the valuable properties, such as easy sampling and tractable density of the original HWN.

We verify the representation learned with RoWN agrees with the common characteristic of representation in hyperbolic space, which is barely observable with the diagonal and the full covariance HWN, by using \emph{synthetic noisy binary tree} dataset.
We demonstrate the usefulness of RoWN on the benchmark datasets: WordNet and Atari 2600 breakout.
We summarize our contributions as follows:
\begin{enumerate}
    \item We provide an analysis of the geometric properties of HWN and its potential limitations in representation learning.
    \item We propose a novel and efficient method of using hyperbolic distribution, namely RoWN, and apply it to probabilistic models.
    \item We demonstrate the performance of RoWN through the comparison with the Euclidean normal distribution, diagonal covariance HWN, and full covariance HWN on one synthetic dataset and two benchmark datasets.
\end{enumerate}

\section{Preliminaries}
This section reviews the wrapped normal distribution defined on the Lorentz model. We first introduce the Lorentz model of hyperbolic space and necessary concepts to understand the wrapped normal distribution.

\subsection{The Lorentz model}
Hyperbolic space is a non-Euclidean space, having a constant negative Gaussian curvature.
\autoref{fig:hyperbolic_space} illustrates two models for hyperbolic space, among four standard equivalent models: (1) Klein model, (2) the Poincar\'e disk model, (3) the Lorentz (hyperboloid) model, and (4) the Poincar\'e half-plane model.
In particular, the Lorentz model is famous for its numerical stability in computing the distance and comes with a simpler closed form of geodesics~\citep{nickel18}.
The Lorentz model $\mathbb{L}^n$ is the Riemannian manifold consisting of the set of points $\rvz \in \mathbb{R}^{n+1}$ satisfying $\langle \rvz, \rvz \rangle_{\mathcal{L}} = -1$ and $z_0 > 0$, where the Lorentizan product $\langle \cdot, \cdot \rangle_{\mathcal{L}}$ is defined as:
\begin{equation*}
    \langle \rvx, \rvy \rangle_{\mathcal{L}} := -x_0y_0 + \sum_{i=1}^n x_iy_i,
\end{equation*}
which also works as the metric tensor on hyperbolic space, i.e., the metric tensor $g$ of the Lorentz model is $g(\boldsymbol{x})= \operatorname{diag}[-1, 1, \cdots, 1]$

\subsection{Tangent space of the Lorentz model}

We denote the tangent space of $\rvx \in \mathbb{L}^n$ as $\mathcal{T}_\rvx \mathbb{L}^n$, which is a set of points satisfying the orthogonality relation with $\rvx$ in terms of the Lorentzian product:
$
T_{\rvx} \mathbb{L}^{n}:=\left\{\boldsymbol{u}:\langle\boldsymbol{u}, \rvx\rangle_{\mathcal{L}}=0\right\}.
$
The metric tensor $g$ induces an inner product of two tangent vectors from a tangent space. 
Geodesic $\gamma:[0, 1]\rightarrow \mathbb{L}^n$ generalizes straight lines in the Riemannian manifold which is the shortest curve between two points. 
The exponential map $\exp_{\rvx}: \mathcal{T}_{\boldsymbol{x}} \mathbb{L}^n \rightarrow \mathbb{L}^n$ maps a tangent vector $\rvu \in \mathcal{T}_\rvx \mathbb{L}^n$ onto $\mathbb{L}^n$ as:
\begin{equation}
    \exp_{\rvx}(\rvu) := \cosh(\Vert \rvu \Vert_\mathcal{L}) \rvx + \sinh(\Vert \rvu \Vert_\mathcal{L}) \frac{\rvu}{\Vert \rvu \Vert_\mathcal{L}},
\end{equation}
such that $\exp_{\rvx}(\rvu)=\rvy, \gamma(0)=\rvx, \gamma(1)=\rvy$.
The log map, inverse of the exponential map, is defined as $\log_\rvx(\rvy) := \exp_{\rvx}^{-1}(\rvu)$.

Parallel transport is an operation that transports a tangent vector in the tangent space at $\rvx$ to another vector in the tangent space at $\rvy$ along the geodesic from $\rvx$ to $\rvy$ without losing the parallel property.
The parallel transport in the Lorentz model is given by:
\begin{equation}
    \mathrm{PT}_{\rvx \rightarrow \rvy}(\rvv) := \rvv + \frac{\langle \rvy - \alpha \rvx, \rvv \rangle_\mathcal{L}}{\alpha + 1}(\rvx + \rvy),
\end{equation}
where $\alpha = -\langle \rvx, \rvy \rangle_\mathcal{L}$.

\subsection{Hyperbolic wrapped normal distribution}
One of the key challenges in adopting hyperbolic space to probabilistic models is finding a distribution on hyperbolic space that is easy to sample and has a closed-form density function.
The two most common distributions of hyperbolic space used in previous work are Riemannian normal distribution~\citep{mathieu19, said14} and hyperbolic wrapped normal distribution (HWN)~\citep{nagano19}.
Our work is mainly based on the HWN because sampling in Riemannian normal distribution is limited with only a unit variance. 

The sampling process with the HWN follows:
\begin{enumerate}
    \item Sample $\rvv \in \R^n$ from a Gaussian distribution $\mathcal{N}(\boldsymbol{0}, \Sigma)$ in Euclidean space.
    \item Parallel transport the vector $[0, \rvv] \in \mathcal{T}_{\boldsymbol{0}_{\mathcal{L}}} \mathbb{L}^n$ to the tangent space.
    \item Project the transported tangent vector to $\mathbb{L}^n$ using exponential mapping.
\end{enumerate}
The density of a sample can be measured via the change of variable method.
For convenience, we denote the procedures 2 and 3 as a single operation:
\begin{equation}
    f_{\boldsymbol{\mu}}(\rvv) :=
    \exp_{\boldsymbol{\mu}}(\mathrm{PT}_{\boldsymbol{0}_{\mathcal{L}} \rightarrow \boldsymbol{\mu}}([0, \rvv])),
    \label{eq:wn_operation}
\end{equation}
with given $\boldsymbol{\mu} \in \mathbb{L}^n$ and $\rvv \in \R^n$, and $\boldsymbol{0}_{\mathcal{L}} = [1, \dots, 0]$ stands for the origin of $\mathbb{L}^n$.

\section{Rotated Hyperbolic Wrapped Normal Distribution}

This section introduces several observations on the geometric properties of the HWN distribution transformation from the Euclidean space to hyperbolic space.
We show the limitations of the diagonal HWN for representation learning of hierarchical data. 
Then, we propose a simple yet effective modification of the HWN, called a rotated hyperbolic wrapped normal distribution.

\subsection{Observations on the hyperbolic wrapped normal distribution}
\label{sec:observations}

\begin{figure}[t!]
    \begin{center}
    \begin{subfigure}[t]{.23\linewidth}
        \begin{tikzpicture}
            \node[anchor=south west] at (0, 0) {\includegraphics[width=\linewidth]{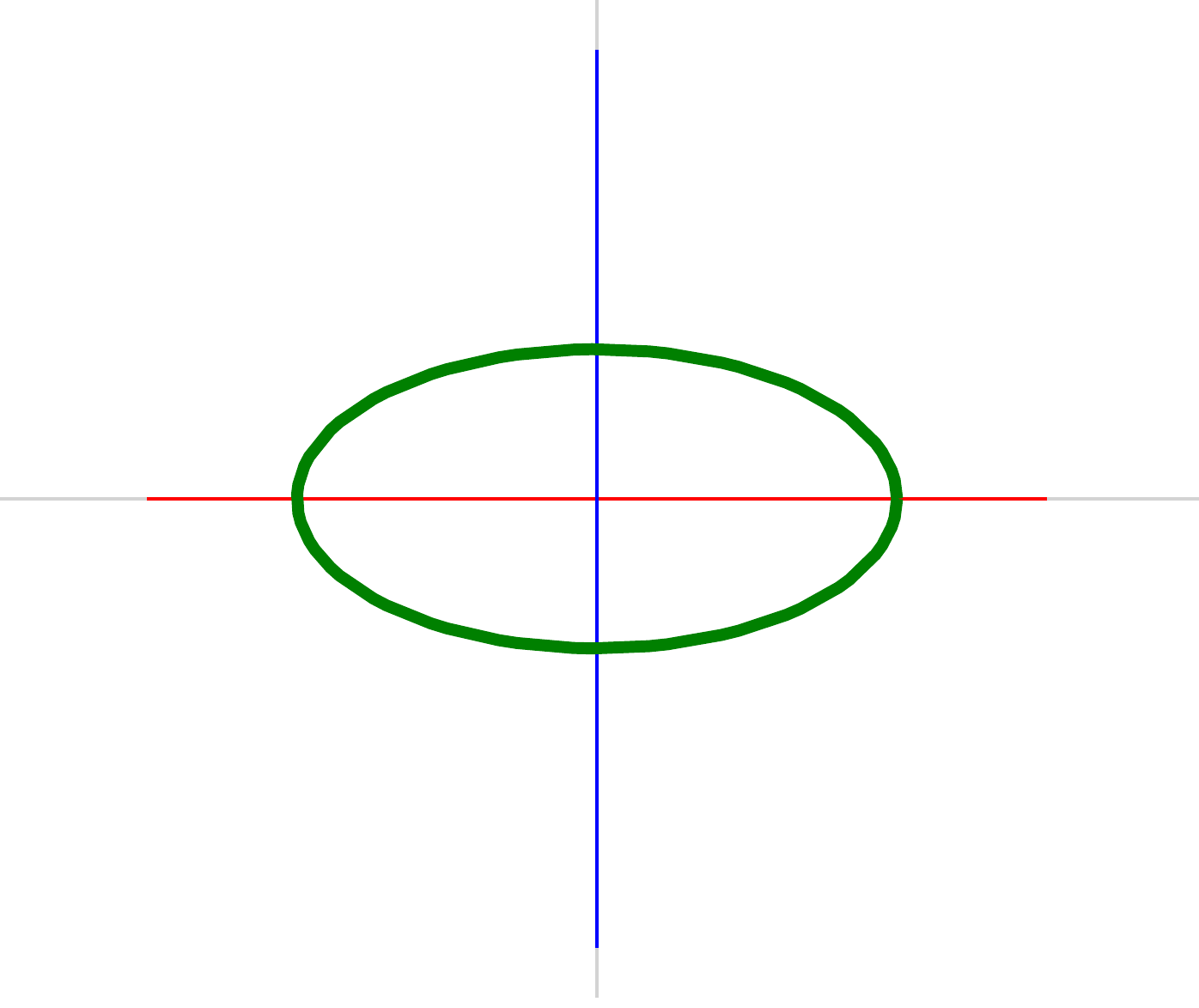}};
            \node[anchor=south west] at (0, 2) {\textsf{\tiny Diagonal HWN}};
        \end{tikzpicture}
        \begin{tikzpicture}
            \node[anchor=south west] at (0, 0) {\includegraphics[width=\linewidth]{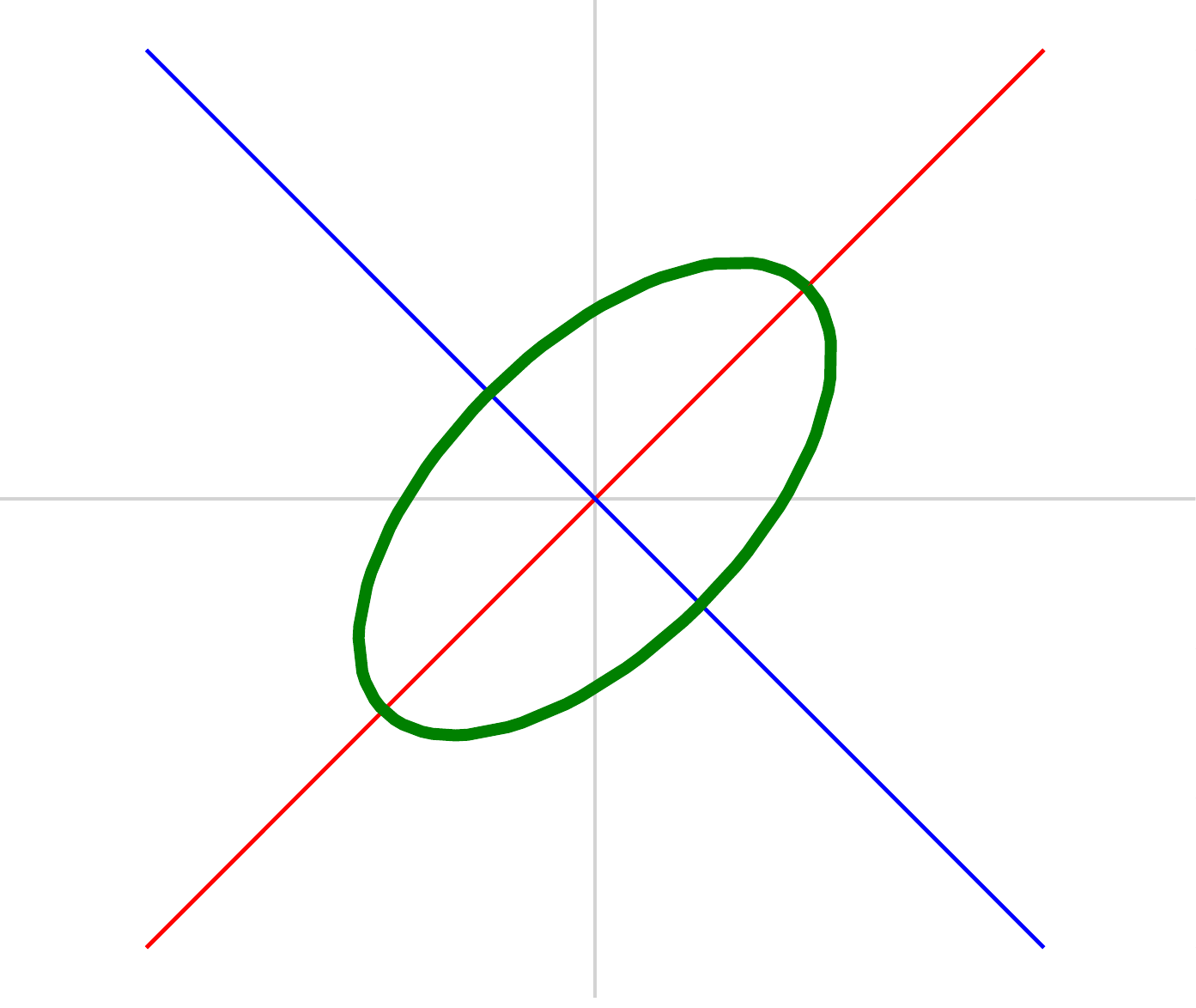}};
            \node[anchor=south west] at (0, 2) {\textsf{\tiny RoWN}};
        \end{tikzpicture}
        \caption{\label{fig:contour_euclidean}
        }
    \end{subfigure}
    \hspace{0.8cm}
    \begin{subfigure}[t]{.23\linewidth}
        \begin{tikzpicture}
            \node[anchor=south west] at (0, 0) {\includegraphics[width=\linewidth]{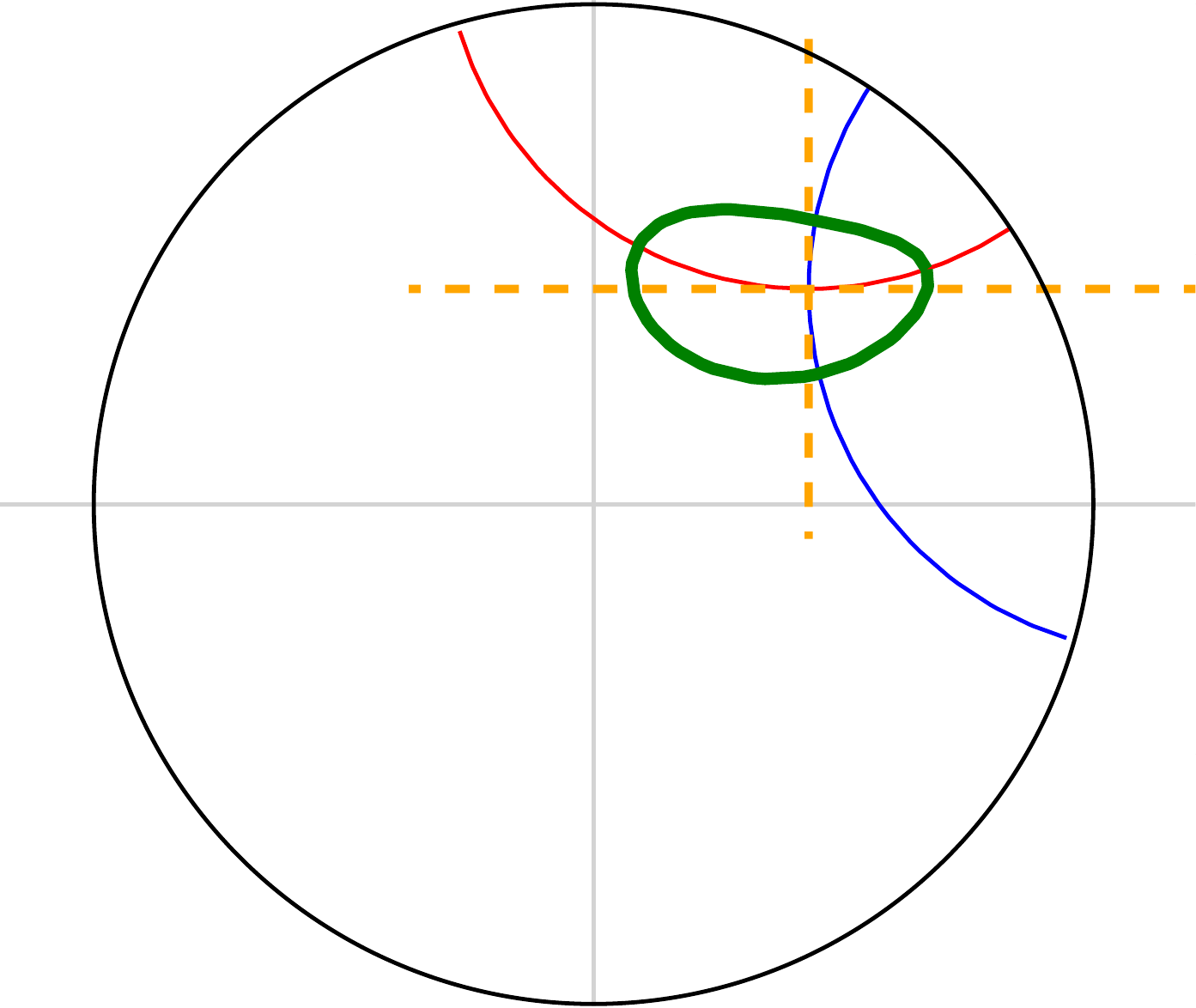}};
        \end{tikzpicture}
        \begin{tikzpicture}
            \node[anchor=south west] at (0, 0) {\includegraphics[width=\linewidth]{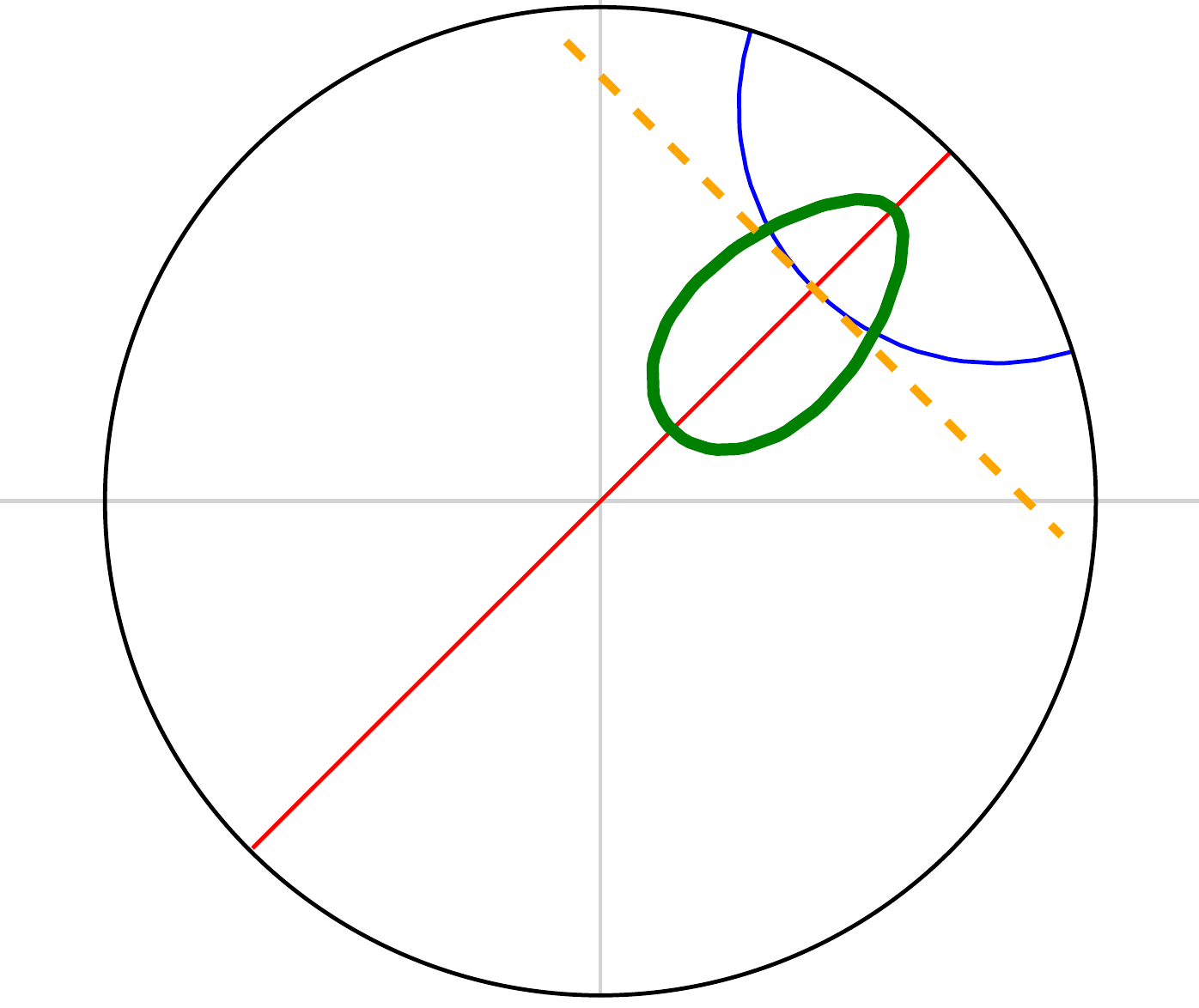}};
        \end{tikzpicture}
        \caption{\label{fig:contour_hyper}}
    \end{subfigure}
    \hspace{1cm}
    \begin{subfigure}[t]{.23\linewidth}
         \begin{tikzpicture}
            \node[anchor=south west] at (0, 0) {\includegraphics[width=\linewidth]{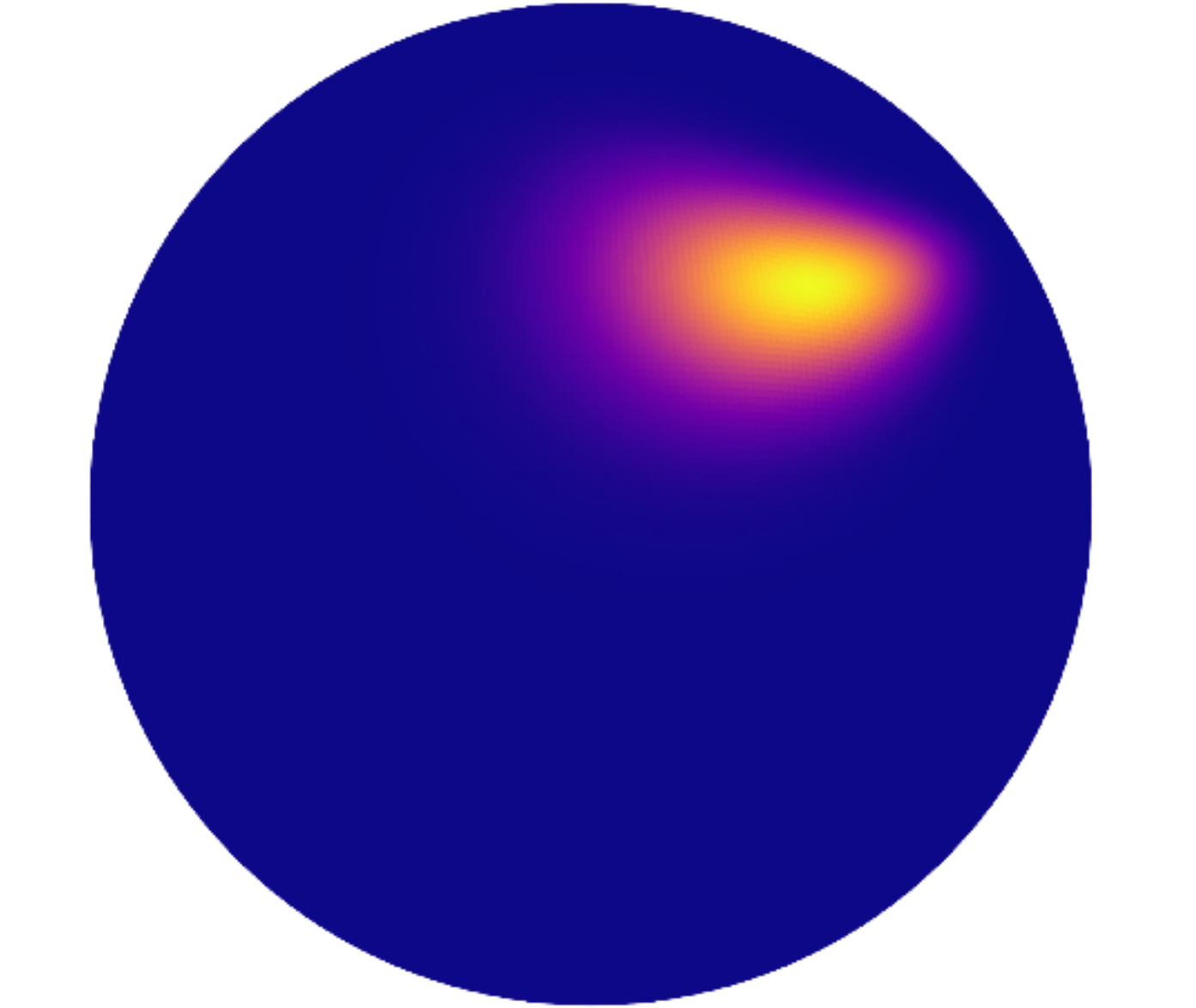}};
        \end{tikzpicture}
        \begin{tikzpicture}
            \node[anchor=south west] at (0, 0) {\includegraphics[width=\linewidth]{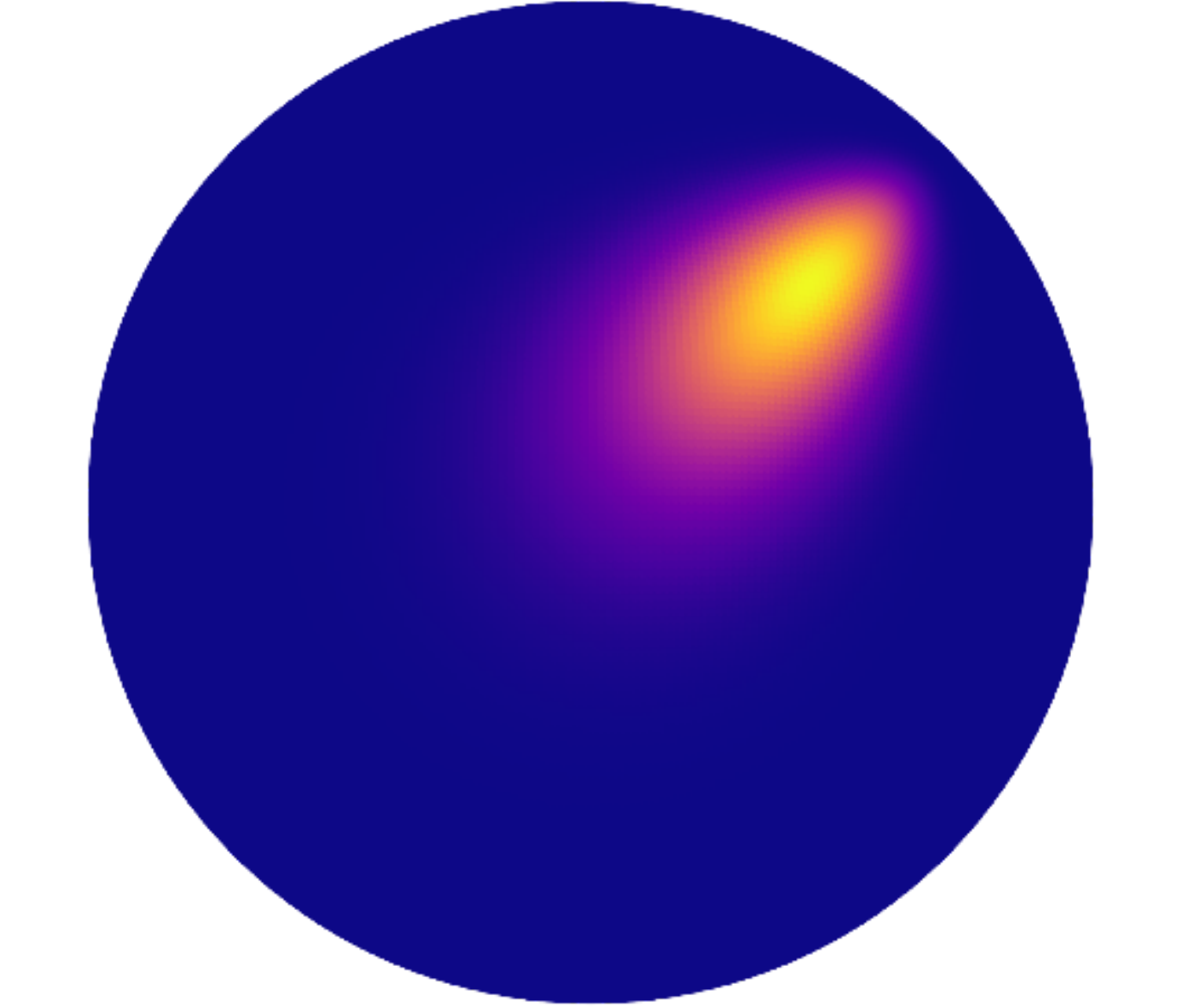}};
        \end{tikzpicture}
        \caption{\label{fig:density_hyper}
        }
    \end{subfigure}
    \caption{
    Visualization of (a) the principal axes of the normal distribution in the Euclidean space, (b) the transported version of two axes in hyperbolic space, and (c) the probability density plot of the distributions.
    (a) The contour line (green) can be represented as an ellipse with major principal axis (red) and minor principal axis (blue).
    (b) The transformed principal axes become geodesics in hyperbolic space.
    The major principal axis of the RoWN passes the hyperbolic origin and crosses with the minor axis at the mean point, whereas the major and minor axes of the diagonal HWN is locally parallel to the standard axes.
    (c) The principal axes determine the shapes of the variational distributions.
    }
    \label{fig:wn_visualization}
    \end{center}
\end{figure}

First, we investigate the changes in the normal distribution during the transformation from Euclidean to hyperbolic space by \autoref{eq:wn_operation}. 
As the principal axes characterize the covariance structure of the normal distribution in the Euclidean space, we investigate how the principal axes are transformed in hyperbolic space.
Before deriving our main proposition, we first show that the straight lines that pass through the origin in the Euclidean space are transformed to the geodesics in hyperbolic space by \autoref{eq:wn_operation}.

\begin{proposition}
\label{proposition:line}
Suppose $\ell_\rvs(t) = t \rvs  \in \R^{n}$ be a line passing through the origin, where $\rvs \in \R^n$ is a directional vector.
Then the curve $f_{\boldsymbol{\mu}}(\ell_\rvs(t))$ in the Lorentz model $\mathbb{L}^n$ becomes a geodesic.
\end{proposition}
The proof of the proposition is provided in \autoref{sec:proof1}. The proposition indicates that every straight line that passes through the origin including the principal axes, is transformed into a geodesic in hyperbolic space.

Based on the first proposition, we provide our main proposition, which fully characterizes the structure of principal axes when projected to the Poincar\'e disk model:
\begin{proposition}
\label{proposition:tangent}
Define $\operatorname{Proj}(\rvu)$ to be the projection function from the Lorentz model to the Poincar\'e model, i.e., $\operatorname{Proj}(\rvu) = \frac{x_{1:}(\rvu)}{x_0(\rvu) + 1},  \forall \rvu \in \mathbb{L}^n$. If $\ell_\rvs$ is a principal axis of the normal distribution defined in $\mathbb{R}^n$ and $\boldsymbol{\mu}$ the mean of HWN in $\mathbb{L}^n$, then $\rvs$ is the tangent vector of $\operatorname{Proj}(f_{\boldsymbol{\mu}}(\ell_\rvs))$ on $\operatorname{Proj}(\boldsymbol{\mu})$.
\end{proposition}
The proof of the proposition is provided in \autoref{sec:proof2}. The proposition reveals that the principal axes of the HWN are locally parallel to the standard bases in the Poincar\'e disk model. To visualize the proposition, we plot the contour line and the principal axes of the two-dimensional diagonal normal distribution before and after the transformation in \autoref{fig:wn_visualization}. 
We observe that the tangent lines of the transformed principal axes are parallel to the standard bases in hyperbolic space. 

When a popular diagonal normal distribution is employed as a variational distribution, the locally parallel principal axes might be problematic in learning hierarchical representations. For example, suppose one tries to represent the variability along the radial direction or in angular differences. In the case, both the major (red) and minor (blue) axes in \autoref{fig:contour_hyper} cannot model the variability properly.

\subsection{Rotated hyperbolic wrapped normal distribution}
Based on the observation, we propose a simple yet effective alternative to the diagonal HWN, a rotated hyperbolic wrapped normal distribution (RoWN).
Rotating the covariance matrix to the direction of $\boldsymbol{\mu}$ enables aligning the major axis of the normal distribution in the Euclidean space to the radial axis in hyperbolic space as visualized in the \autoref{fig:wn_visualization}. 

\begin{algorithm}[t!]
\setstretch{1.3}
\caption{Sampling process with the rotated hyperbolic wrapped normal distribution}
 \textbf{Input} Mean $\boldsymbol{\mu} \in \mathbb{L}^n$, diagonal covariance matrix $\Sigma \in \R^{n \times n}$ \\
 \textbf{Output} Sample $\rvz \in \mathbb{L}^n$
\begin{algorithmic}[1]
\State $\rvx = [\pm 1, \dots, 0] \in \R^n, \rvy = \boldsymbol{\mu}_{1:} / \Vert \boldsymbol{\mu}_{1:} \Vert$ \Comment{$\pm$ is determined by the sign of $\boldsymbol{\mu}_0$}
\State $\mR = \mI + (\rvy^T \rvx - \rvx^T \rvy) + (\rvy^T \rvx - \rvx^T \rvy)^2 / (1 + \langle \rvx, \rvy \rangle)$
\State Rotate $\hat{\Sigma} = \mR \Sigma \mR^T$
\State Sample $\rvv \sim \mathcal{N}(\boldsymbol{0}, \hat{\Sigma})$
\State \textbf{return} $\rvz = f_{\boldsymbol{\mu}}(\rvv)$
\end{algorithmic}
\label{alg:rown_sampling_main}
\end{algorithm}

\begin{figure}[b!]
    \centering
    \begin{subfigure}[t]{\dimexpr0.23\linewidth+30pt\relax}
        \makebox[35pt]{\raisebox{50pt}{\textsf{\tiny samples}}}%
        \includegraphics[width=\dimexpr\linewidth-30pt\relax]{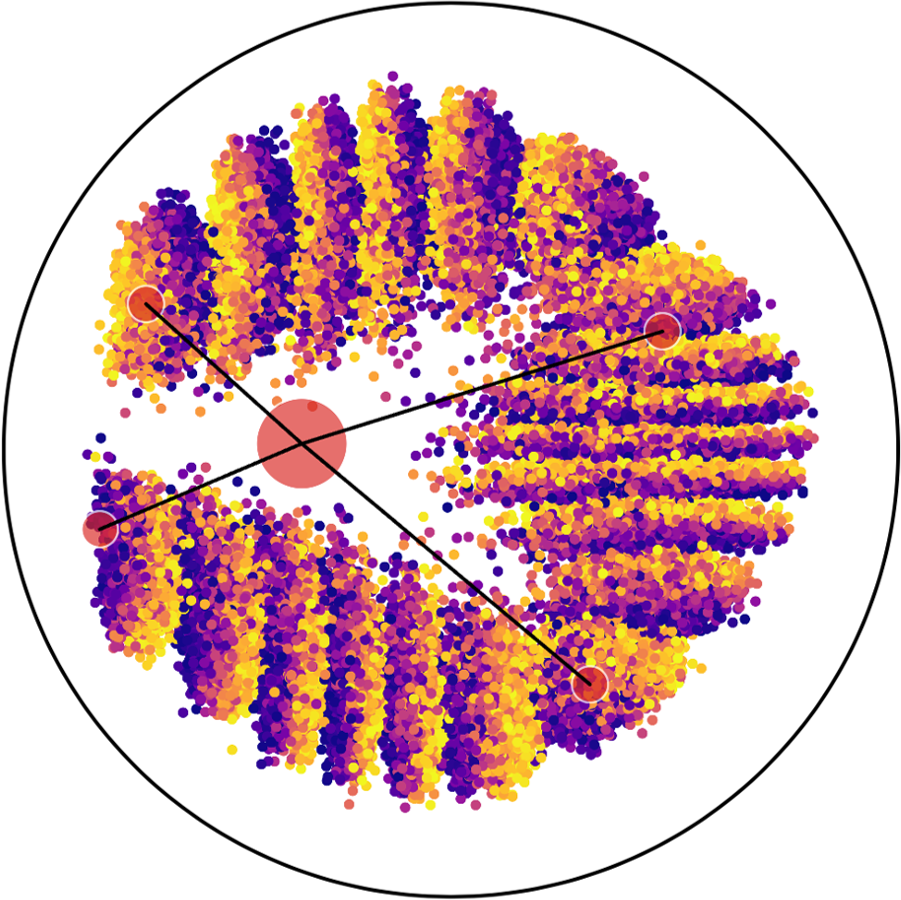}
        \makebox[35pt]{\raisebox{45pt}{\textsf{\tiny means}}}%
        \includegraphics[width=\dimexpr\linewidth-30pt\relax]{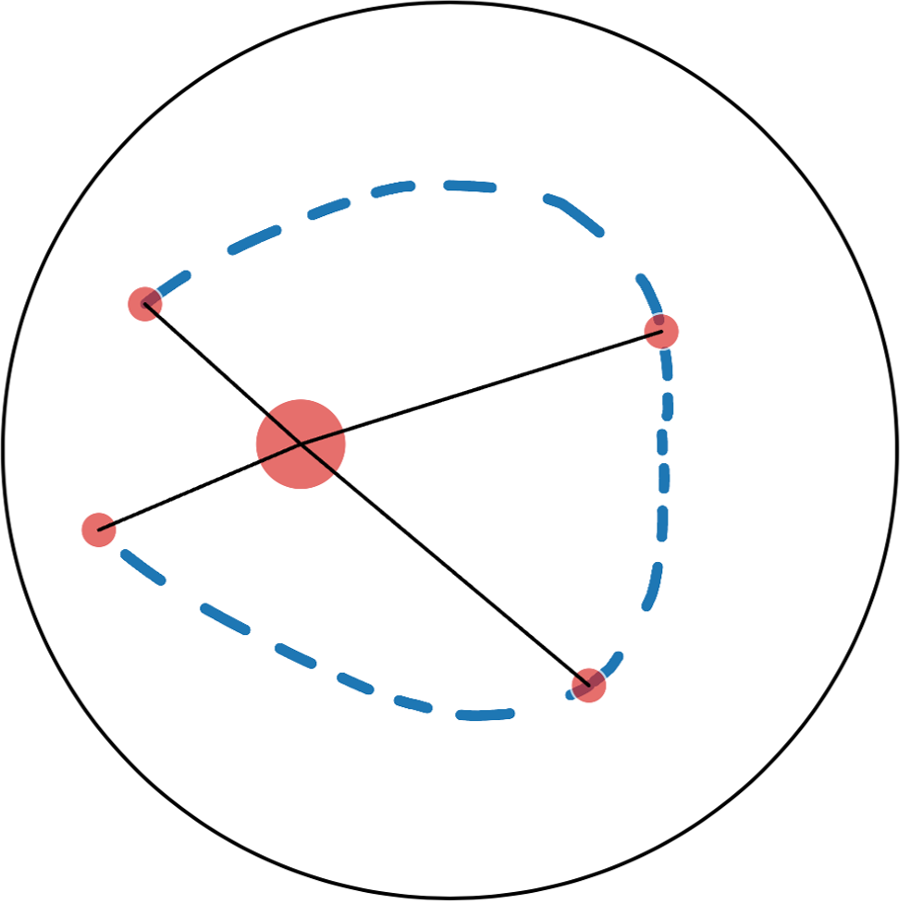}
        \caption{\label{fig:hwn_plots}Diagonal HWN}
    \end{subfigure}
    \hspace{1cm}
    \begin{subfigure}[t]{.23\linewidth}
        \includegraphics[width=\linewidth]{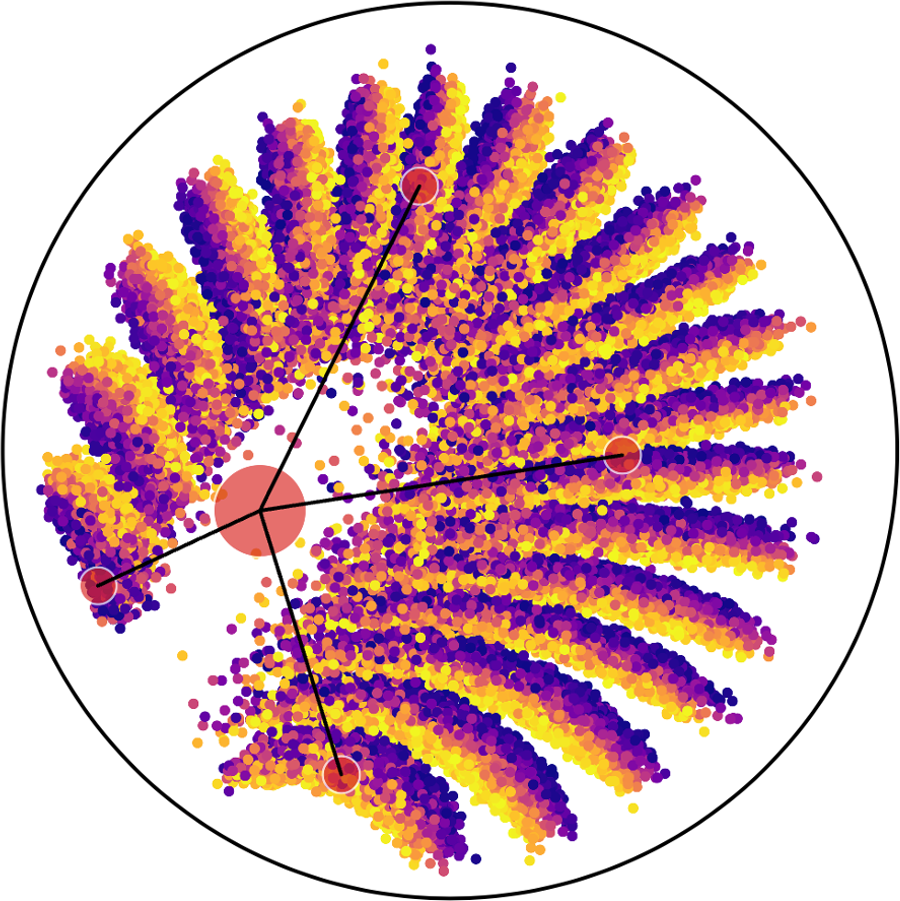}
        \includegraphics[width=\linewidth]{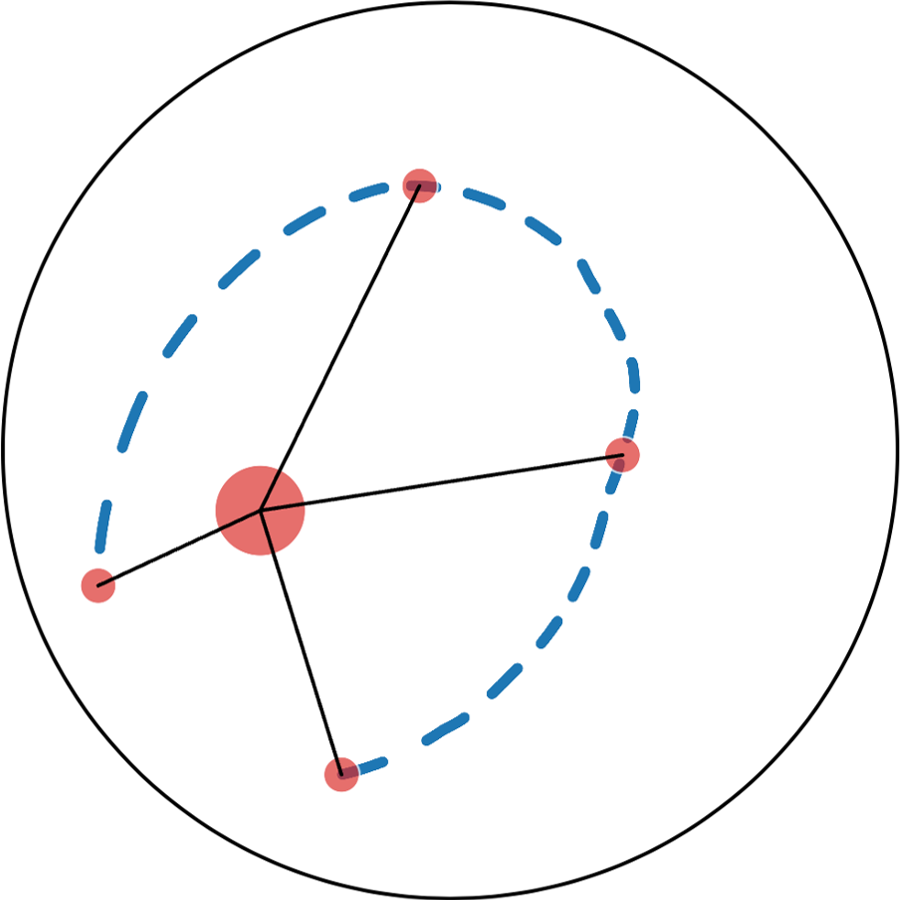}
        \caption{\label{fig:full_hwn_plots}Full covariance HWN}
    \end{subfigure}
    \hspace{0.8cm}
    \begin{subfigure}[t]{.23\linewidth}
        \includegraphics[width=\linewidth]{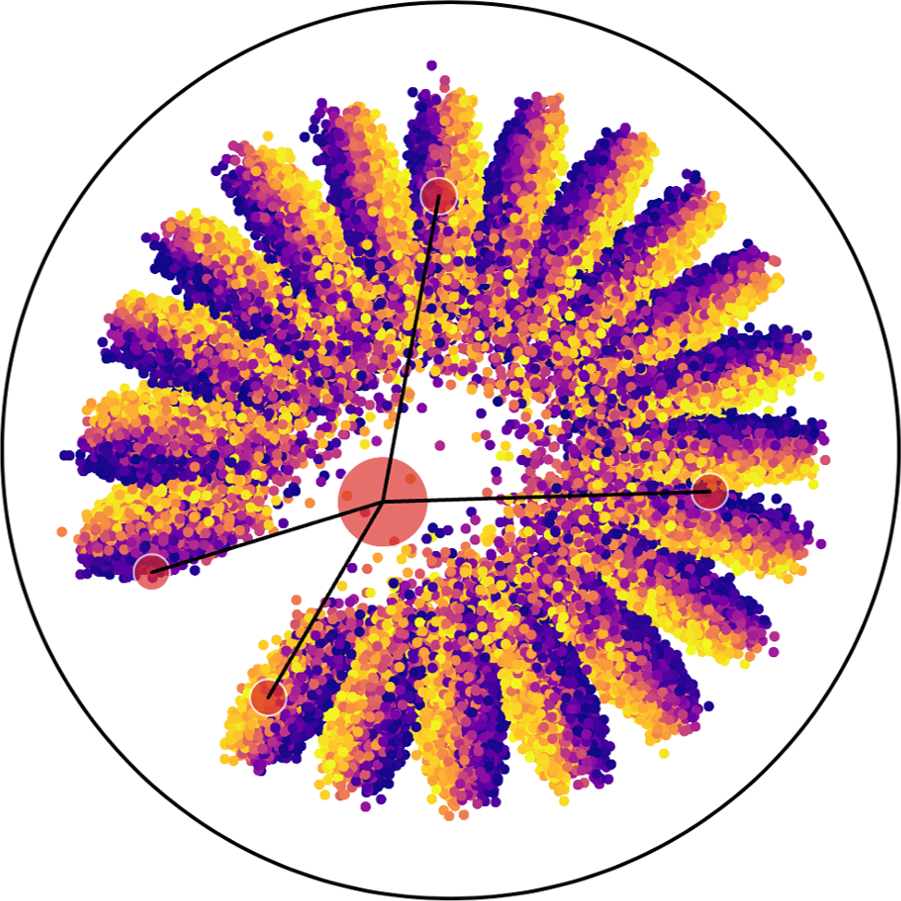}
        \includegraphics[width=\linewidth]{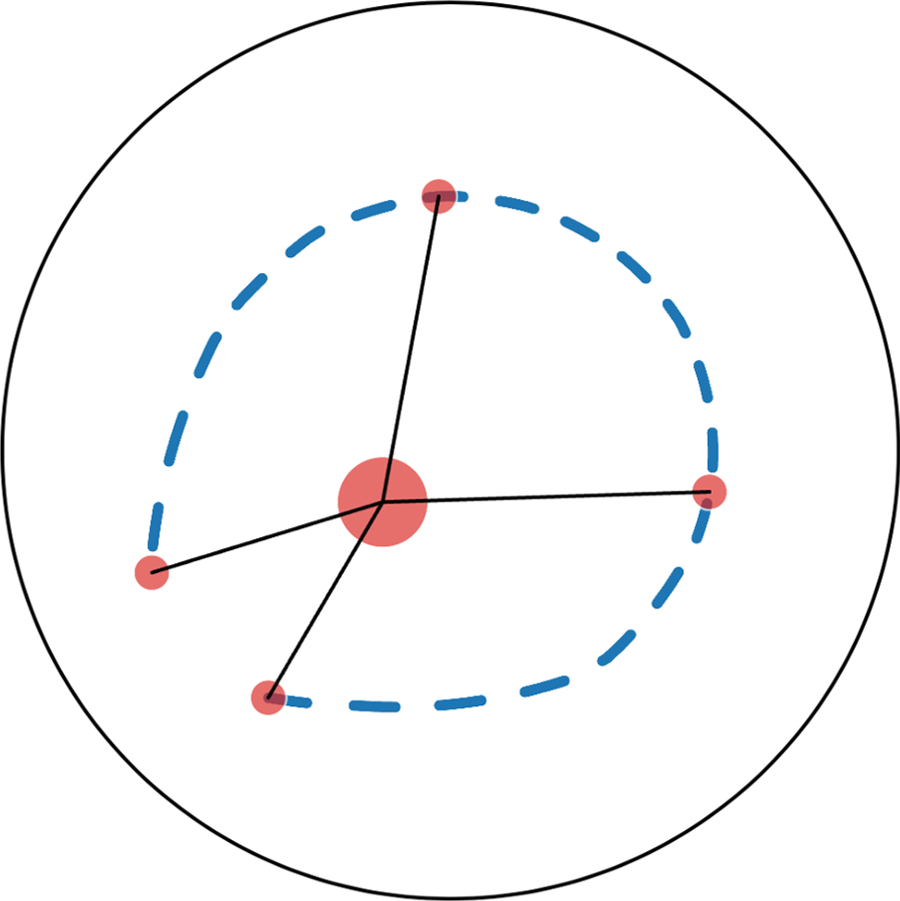}
        \caption{\label{fig:rown_plots}RoWN}
    \end{subfigure}
    \caption{
    Visualization of the variational distribution of hyperbolic VAE on a synthetic binary tree dataset with different variational distributions.
    The red dots denote the variational means of the root and four representative children. 
    The upper row shows the samples from the variational distributions where the color denotes the level of noise. 
    The bottom row shows the means of the variational distributions.
    Overall, RoWN better aligns local variation in angular difference.
    }
    \label{fig:WN_failure}
\end{figure}

To construct the distribution, we start with a mean vector $\boldsymbol{\mu} \in \mathbb{L}^n$ and a diagonal covariance matrix $\Sigma$ as in the standard HWN.
We change the covariance matrix of the normal distribution as follows:
\begin{enumerate}[topsep=0pt]
    \item Compute the rotation matrix $\mR$ that rotates the x-axis ($[\pm1, \dots, 0] \in \R^n$) to $\boldsymbol{\mu}_{1:}$.
    \item Substitute the covariance matrix of Gaussian normal with $\mR \Sigma \mR^T$.
\end{enumerate}
Thus, the rotation matrix $\mR$, which rotates a unit vector from $\rvx$ to $\rvy$, can be computed as:
\begin{equation}
    \label{eq:rotation}
    \mR = \mI + (\rvy^T \rvx - \rvx^T \rvy) + \frac{1}{1 + \langle \rvx, \rvy \rangle} (\rvy^T \rvx - \rvx^T \rvy)^2.
\end{equation}

The pseudo-code of the sampling process of RoWN is in \autoref{alg:rown_sampling_main}.
Note that the construction is straightforward but still keeps the following benefits of the HWN:
1) The sampling can be done efficiently, and 2) the computation of the probability density of the samples is tractable.
As the HWN provides a tractable probability density function for any kind of covariance matrix, we can easily compute the probability density of a given sample from RoWN. {See \Apxref{sec:rown_apx} for more details.}

We empirically demonstrate the influence of different variational distributions in \autoref{fig:WN_failure}. We visualize the variational distributions of {the} synthetic binary tree dataset of depth two after training a hyperbolic VAE. 
As the result shows, the model with the diagonal HWN represents the variation in a child parallel to the standard bases, whereas the model with RoWN represents the variation in angular difference.
Please check the detailed description on the synthetic dataset in \autoref{sec:noisy_binary_tree}.

\section{Experiments}
\label{sec:experiments}
In this section, we first explain the two applications of the distribution defined on hyperbolic space: hyperbolic VAE and probabilistic hyperbolic word embedding model.
We then conduct three different experiments to compare the performance of RoWN with four baselines, including the Gaussian distribution in the Euclidean space, the isotropic HWN~\citep{nagano19}, the diagonal HWN, and the full covariance HWN.
We also provide an additional study on a variant of RoWN with learnable rotation direction $\rvy$ in \autoref{alg:rown_sampling_main}, and the results are in \autoref{tab:nbt_results_learnable_x_apx}.
The details of the experiments are described in \autoref{sec:experiments_apx}.

\subsection{Applications of the hyperbolic distribution}

\paragraph{Hyperbolic VAE.}
The hyperbolic VAE, whose latent space is hyperbolic space, has been shown to be efficient for capturing the hierarchical structure of the data~\citep{nagano19,mathieu19}.
The evidence lower bound of the hyperbolic VAE can be written as:
\begin{equation*}
    \mathcal{L}_{\mathrm{ELBO}}(\theta, \phi) := \E_{q_\phi(\rvz \mid \rvx) \cdot \sqrt{\det(g)}}[\log p_\theta(\rvx  \mid \rvz)] - \KL\left(q_\phi(\rvx \mid \rvz) \cdot \sqrt{\det(g)} \parallel p(\rvz) \cdot \sqrt{\det(g)}\right),
\end{equation*}
where $q_\phi$ is the encoder, $p_\theta$ is the decoder, $g$ is the metric tensor of the chosen model of hyperbolic space, and $p(\rvz)$ is a prior distribution.
The distributions defined on hyperbolic spaces, such as HWN and  RoWN, are used to define encoder $q_\phi(\rvz \mid \rvx) \cdot \sqrt{\det(g)}$ and prior $p(\rvz) \cdot \sqrt{\det(g)}$.
In hyperbolic VAEs, due to the absence of the closed-form KL divergence in the hyperbolic distributions, the KL divergence between the encoder and the prior is usually approximated with Monte-Carlo sampling~\citep{nagano19,mathieu19}.

\paragraph{Probabilistic hyperbolic word embedding model.}
The probabilistic word embedding models aim to learn probabilistic representations of words~\citep{vilnis15,nagano19,tifrea19}.
The embeddings learned on hyperbolic spaces have shown better performance than the Euclidean counterpart~\citep{nagano19,tifrea19}.
Given the hypernymy relationships between the words, the probabilistic hyperbolic embedding model learns the probabilistic representation of words by minimizing the following objective:
\begin{equation}
    \mathcal{L}_{\mathrm{word}}(\boldsymbol{\theta}) := \E_{(s \sim t, s \not\sim t')}[\max(0, m + \KL(q_s \parallel q_t) - \KL(q_s \parallel q_{t'}))],
\end{equation}
where $m$ is a margin, $q_i$ is a distribution for word $i$ parameterized by $\theta$, and $s \sim t$ and $s \not\sim t$ denote the presence and absence of hypernymy relation between word pair $s$ and $t$, respectively.

\subsection{Noisy synthetic binary tree}
\label{sec:noisy_binary_tree}

\begin{figure}[t!]
    \centering
    \begin{subfigure}[t]{.45\linewidth}
        \centering
        \includegraphics[width=1.2\linewidth]{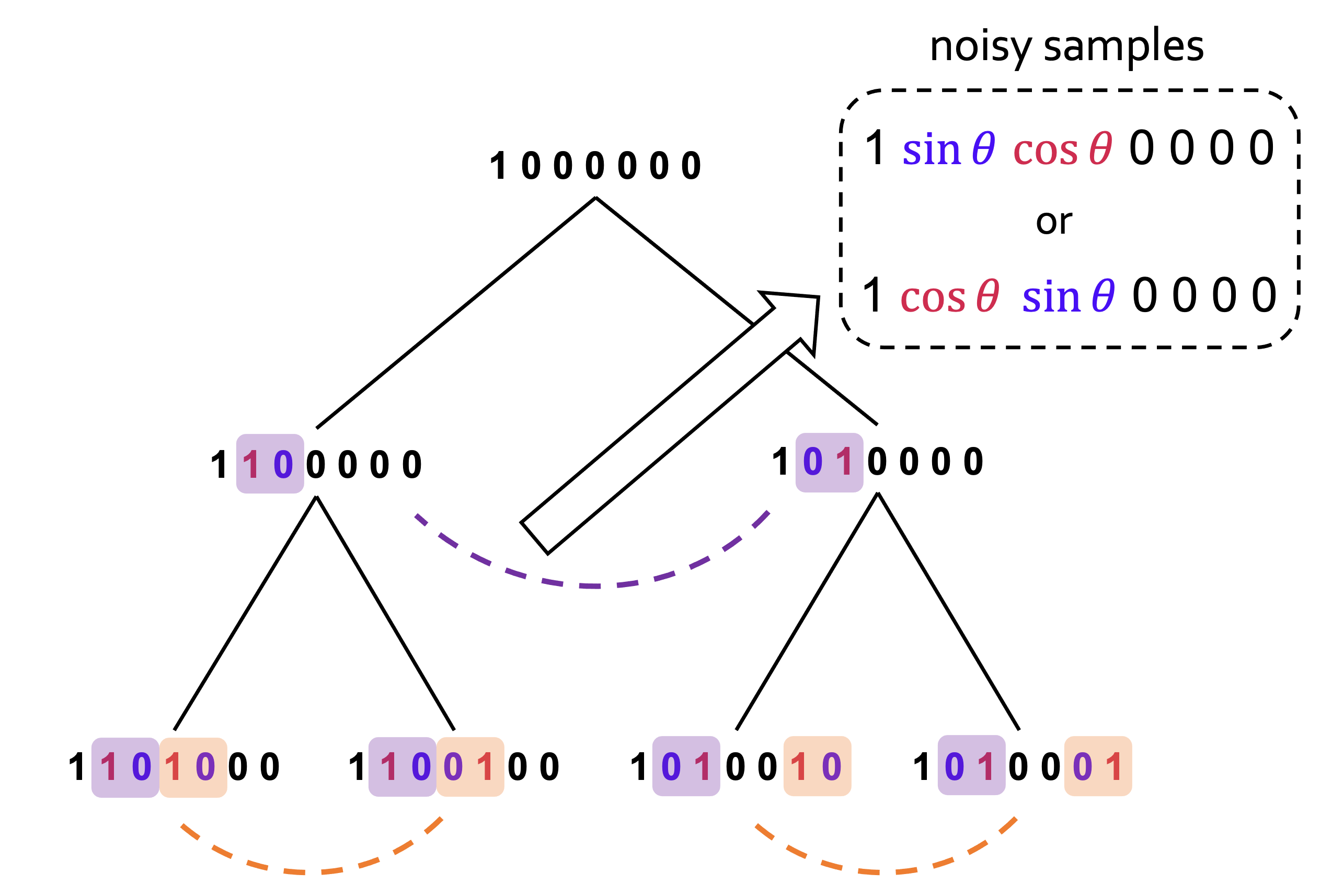}
        \caption{Noisy synthetic binary tree\label{fig:syn_three_sub}}
    \end{subfigure}
    \hspace{10mm}
    \begin{subfigure}[t]{.45\linewidth}
        \centering
        \includegraphics[width=0.8\linewidth]{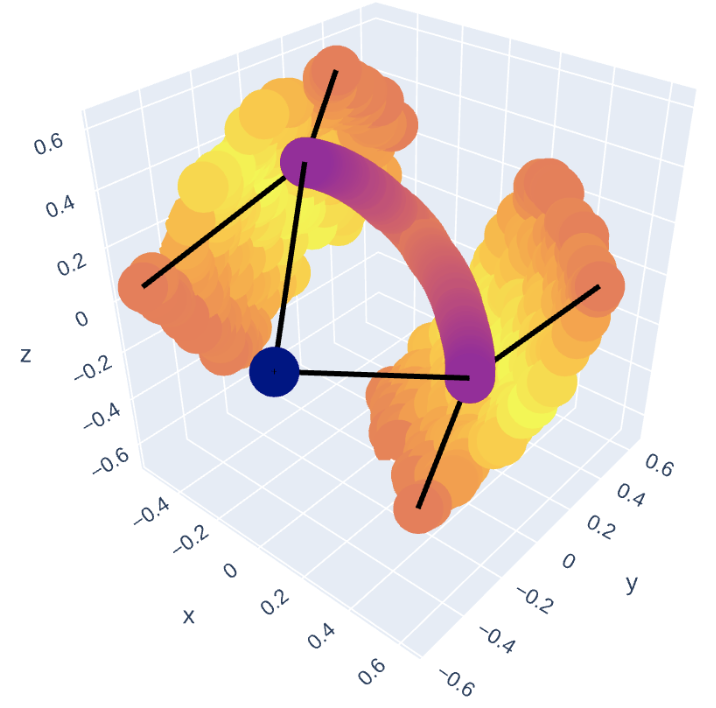}
        \caption{Learned representations\label{fig:syn_viz}}
    \end{subfigure}
    \hfill
    \caption{Illustration of a noisy synthetic binary tree. 
    (a) We construct a noisy synthetic binary tree by adding spherical noises defined with $\theta$.
    The continuous samples are generated at the same distance from the root. 
    (b) We train the depth three noisy synthetic binary tree with hyperbolic VAE with RoWN as a variational distribution and visualize the means of the variational distributions. 
    The black lines show the underlying hierarchical structure, and the color denotes the level of noise. 
    The hierarchy and the local variations are well preserved in the representations.}
    \label{fig:nbt_example}
\end{figure}

A synthetic binary tree dataset is first used to show the performance of representing hierarchy in \citet{nagano19}, where each node in a tree corresponds to a sequence of binary values. 
\autoref{fig:syn_three_sub} shows an example of the depth three binary tree, where a parent and child only differ in one digit.
We add spherical noises to the nodes in the same level of hierarchy as described in \autoref{fig:syn_three_sub} as the noisy samples. With the noisy samples, we can create a dataset containing a local-level variation in the hierarchy. For the experiments, we uniformly sample the spherical noise from  $[0, \pi/4]$.

\begin{table}[b!]
    \centering
    \caption{Results of \textit{noisy synthetic binary tree}. 
    The results are averaged over 10 runs.
    The hyperbolic models outperform the Euclidean model in all settings.
    Overall, RoWN preserves the hierarchical information better than the other distributions.
    }
    \vspace{1em}
    \label{tab:nbt_results}
    \begin{tabular}{c l c c c c}
        \toprule
         & & \multicolumn{4}{c}{Depth} \\
         \cmidrule{3-6}
         & & 4 & 5 & 6 & 7 \\
         \midrule
         \multirow{5}{*}{\shortstack[c]{Correlation\\w/ distance}} 
         & Euclidean & $0.748_{\pm .032}$ & $0.740_{\pm .013}$ & $0.741_{\pm .008}$ & $0.733_{\pm .014}$ \\
         & HWN (isotropic $\Sigma$) & $0.773_{\pm .030}$ & \cellcolor{yellow!70}$0.809_{\pm .016}$ & $0.798_{\pm .008}$ & $0.735_{\pm .022}$ \\
         & HWN (diagonal $\Sigma)$ & $0.814_{\pm .008}$ & $0.791_{\pm .023}$ & $0.817_{\pm .010}$ & $0.759_{\pm .025}$ \\
         & HWN (full $\Sigma$) & \cellcolor{yellow!70}$0.827_{\pm .015}$ & $0.798_{\pm .026}$ & $0.798_{\pm .010}$ & \cellcolor{yellow!70}$0.794_{\pm .014}$ \\
         & RoWN & $0.820_{\pm .015}$ & $0.807_{\pm .017}$ & \cellcolor{yellow!70}$0.822_{\pm .017}$ & $0.788_{\pm .016}$  \\
         \midrule
         \multirow{5}{*}{\shortstack[c]{Correlation\\w/ depth}} 
         & Euclidean & $0.762_{\pm .117}$ & $0.807_{\pm .038}$ & $0.712_{\pm .054}$ & $0.612_{\pm .049}$ \\
         & HWN (isotropic $\Sigma$) & $0.902_{\pm .033}$ & $0.867_{\pm .034}$ & $0.811_{\pm .029}$ & $0.602_{\pm .066}$ \\
         & HWN (diagonal $\Sigma$) & $0.918_{\pm .028}$ & $0.808_{\pm .076}$ & $0.862_{\pm .035}$ & $0.697_{\pm .076}$ \\
         & HWN (full $\Sigma$) & \cellcolor{yellow!70}$0.956_{\pm .015}$ & $0.878_{\pm .051}$ & $0.870_{\pm .033}$ & $0.815_{\pm .055}$ \\
         & RoWN & $0.930_{\pm .026}$ & \cellcolor{yellow!70}$0.911_{\pm .027}$ & \cellcolor{yellow!70}$0.901_{\pm .034}$ & \cellcolor{yellow!70}$0.827_{\pm .047}$ \\
         \bottomrule
    \end{tabular}
\end{table}

We train hyperbolic VAE on \textit{noisy synthetic binary tree} with varying depths. The detailed model description is available in \Apxref{sec:baselines}. We set the latent dimension the same as the depth.
We report 1) the correlation between the hamming distance and the embedding distance and 2) the correlation between the depth and the Poincar\'e norm of the embeddings. The first correlation is computed over all possible pairs of test points.
As \autoref{tab:nbt_results} shows, the full covariance HWN and RoWN improve the diagonal HWN except depth six, outperforming the Euclidean model in every setting. RoWN preserves the depth information better than the other distributions in general.
We additionally visualize the variational mean obtained by training the tree of depth three in \autoref{fig:syn_viz}, where the hierarchical structure is well preserved in the hyperbolic embedding space.

\subsection{WordNet}

\begin{table}[t!]
    \centering
    \caption{Results of \textit{WordNet}. 
    The results are an average of 5 runs.
    Based on the rank of hypernymy pairs among non-hypernymy pairs, we report the mean rank (MR) and mean average precision (mAP) for evaluation.}
    \label{tab:wordnet_results}
    \vspace{1em}
    \begin{tabular}{c l c c c}
        \toprule
         & & \multicolumn{3}{c}{Latent dimension} \\
         \cmidrule{3-5}
         &  & 5 & 10 & 20 \\
         \midrule
         \multirow{5}{*}{MR} 
         & Euclidean & \cellcolor{yellow!70}$13.968_{\pm 0.504}$ & $3.862_{\pm 0.281}$ & $1.955_{\pm 0.157}$ \\
         & HWN (isotropic $\Sigma$) & $14.568_{\pm 2.203}$ & $4.470_{\pm 0.669}$ & $3.125_{\pm 0.455}$\\
         & HWN (diagonal $\Sigma$) & $16.590_{\pm 1.146}$ & $3.891_{\pm 0.447}$ & $2.062_{\pm 0.088}$\\
         & HWN (full $\Sigma$) & $557.309_{\pm 18.006}$ & $466.513_{\pm 75.142}$ & $599.140_{\pm 18.916}$ \\
         & RoWN & $16.271_{\pm 2.985}$ & \cellcolor{yellow!70}$2.888_{\pm 0.162}$ & \cellcolor{yellow!70}$1.783_{\pm 0.090}$\\
         \midrule
         \multirow{5}{*}{mAP} 
         & Euclidean & $0.565_{\pm 0.014}$ & $0.801_{\pm 0.020}$ & $0.902_{\pm 0.008}$\\
         & HWN (isotropic $\Sigma$) & \cellcolor{yellow!70}$0.617_{\pm 0.012}$ & $0.820_{\pm 0.013}$ & $0.847_{\pm 0.017}$\\
         & HWN (diagonal $\Sigma$)& $0.565_{\pm 0.020}$ & $0.805_{\pm 0.015}$ & $0.905_{\pm 0.007}$\\
         & HWN (full $\Sigma$) & $0.032_{\pm 0.003}$ & $0.063_{\pm 0.005}$ & $0.079_{\pm 0.021}$ \\
         & RoWN & $0.593_{\pm 0.024}$ & \cellcolor{yellow!70}$0.844_{\pm 0.009}$ & \cellcolor{yellow!70}$0.921_{\pm 0.005}$\\
         \bottomrule
    \end{tabular}
\end{table}

We train a probabilistic word embedding model with WordNet dataset~\citep{fellbaum98}, which consists of 82,115 nouns and 743,241 hypernymy relationships.
We have initialized the embeddings from $\mathcal{N}(0, 0.01 I)$, which are then moved to the Lorentz model using the exponential map. We use the learning rate warm-up proposed in \citep{nagano19}.
We evaluate the learned representations by computing the average rank of all the hypernymies. 
The rank of a given pair of words $s$ and $t$ is computed among the distances between all possible pairs of the words $s$ and $t'$ without hypernymy.
\autoref{tab:wordnet_results} shows the empirical performances of representing the word data. We report the performance with the mean rank (MR) and the mean average precision (mAP).
RoWN preserves the hierarchical structure better than the other distributions, while the full covariance HWN often performs worse than RoWN.

\paragraph{Failure of the full covariance HWN.}
In theory, the full covariance HWN needs to have at least a similar performance to RoWN since the RoWN is the particular case of the full covariance.
The performance of the full covariance HWN often performs worse with the probabilistic hyperbolic word embedding models than the hyperbolic VAEs. 
We speculate that reason is because of the relatively simple prior in the hyperbolic VAE, whereas the probabilistic word embedding models need to compute the KL divergence between the full covariance HWNs.
Our additional experiments reported in \autoref{tab:wordnet_train_samples} confirm our speculation as the number of training samples for the KL divergence increases, the performance increases slightly. 

\paragraph{Discussion of the root placement.}
Note that due to the isometry of the geometry, the root node can be placed anywhere in hyperbolic space. 
In other words, infinitely many sets of embeddings preserve the same pairwise distances between the nodes.
To place the root near the origin, we initialize all embeddings with the zero mean distribution as done in \citep{nagano19}.
We find that this initialization helps the root placed near the origin.
A study on the effects of the initialization methods is shown in \autoref{tab:root_node_results}.
Further details about the root placement can be found in \autoref{sec:root_placement}.

\begin{table}[b!]
    \centering
    \caption{
    The effects of initializations. 
    We test different initializations on the deterministic hyperbolic word embedding models trained with the subset of the WordNet dataset.
    The \texttt{near zero vector} model initializes the embeddings with uniform distribution $\mathcal{U}(-0.001, 0.001)$, while the \texttt{near one vector} model initializes the embeddings with uniform distribution $\mathcal{U}(0.999, 1.001)$.
    While the Poincar\'e norm of the root node differs, the other metrics remain similar.
    }
    \label{tab:root_node_results}
    \begin{tabular}{c l c c c c}
        \toprule
         & & \multicolumn{4}{c}{Latent dimension} \\
         \cmidrule{3-6}
         & & 2 & 5 & 10 & 20 \\
         \midrule
         \multirow{2}{*}{\shortstack[c]{MR}} 
         & Near zero vector & $4.346_{\pm .643}$ & $3.270_{\pm .144}$ & $2.828_{\pm .098}$ & $2.508_{\pm .049}$ \\
         & Near one vector & $4.029_{\pm .415}$ & $3.209_{\pm .094}$ & $2.856_{\pm .075}$ & $2.491_{\pm .051}$ \\
         \midrule
         \multirow{2}{*}{\shortstack[c]{mAP}} 
         & Near zero vector & $0.821_{\pm .016}$ & $0.891_{\pm .006}$ & $0.891_{\pm .005}$ & $0.895_{\pm .003}$ \\
         & Near one vector & $0.829_{\pm .010}$ & $0.894_{\pm .004}$ & $0.891_{\pm .004}$ & $0.896_{\pm .003}$ \\
         \midrule
         \multirow{2}{*}{\shortstack[c]{The Poincar\'e norm of \\the root node}} 
         & Near zero vector & \cellcolor{yellow!70}$0.122_{\pm .036}$ & \cellcolor{yellow!70}$0.042_{\pm .015}$ & \cellcolor{yellow!70}$0.031_{\pm .007}$ & \cellcolor{yellow!70}$0.024_{\pm .004}$ \\
         & Near one vector & $0.505_{\pm .076}$ & $0.650_{\pm .010}$ & $0.732_{\pm .003}$ & $0.801_{\pm .001}$ \\
         \bottomrule
    \end{tabular}
\end{table}

\subsection{Atari 2600 Breakout}

\begin{table}[t!]
    \centering
    \caption{Results of \textit{Atari 2600 Breakout}. 
    The results are averaged over 5 runs.
    We measure the correlation between the score of an image and the Poincar\'e norm of the variational mean.
    The HWN with the isotropic covariance can be viewed as a variant of RoWN.}
    \label{tab:atari_results}
    \vspace{1em}
    \begin{tabular}{c l c c c}
        \toprule
         & & \multicolumn{3}{c}{Latent dimension} \\
         \cmidrule{3-5}
         & & 10 & 15 & 20 \\
         \midrule
         \multirow{5}{*}{\shortstack[c]{Correlation btw.\\score and norm}} 
         & Euclidean & $0.379_{\pm .007}$ & $0.436_{\pm .029}$ & $0.479_{\pm .020}$\\
         & HWN (isotropic $\Sigma$) & \cellcolor{yellow!70}$0.513_{\pm .012}$ & \cellcolor{yellow!70}$0.598_{\pm .021}$ & \cellcolor{yellow!70}$0.607_{\pm .015}$ \\
         & HWN (diagonal $\Sigma$) & $0.478_{\pm .011}$ & $0.513_{\pm .006}$ & $0.513_{\pm .008}$ \\
         & HWN (full $\Sigma$) & $0.483_{\pm .011}$ & $0.520_{\pm .009}$ & $0.563_{\pm .010}$ \\
         & RoWN & $0.497_{\pm .014}$ & $0.556_{\pm .014}$ & $0.561_{\pm .029}$ \\
         \bottomrule
    \end{tabular}
\end{table}

Trajectories of some Atari 2600 games can be structured as a tree-like hierarchy along the time horizon as \citet{nagano19} points out.
For example, given Atari 2600 Breakout, the root node can be the starting state of the game where no blocks have been broken. As the game progresses the states of the blocks form a hierarchical structure depending on which blocks have been broken.
To learn the implicit hierarchy that can be observed from the trajectories of Breakout, we train the VAE models with the Atari 2600 Breakout images.

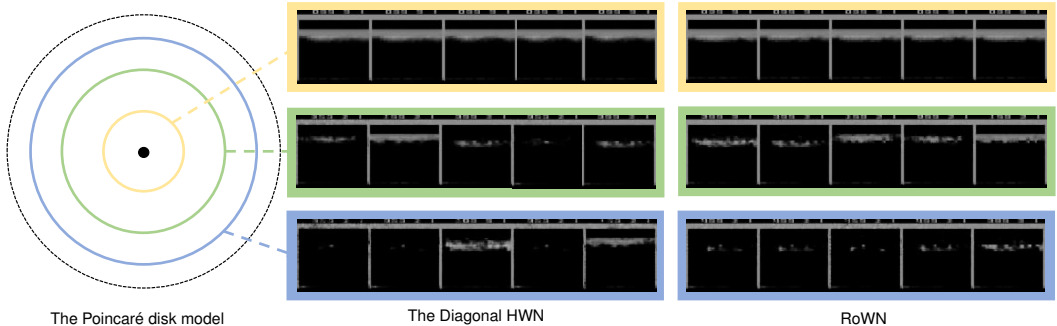
\begin{figure}[b!]
    \centering
        \begin{tikzpicture}
            \node[anchor=south west] at (0, 0.3) {\includegraphics[width=\linewidth]{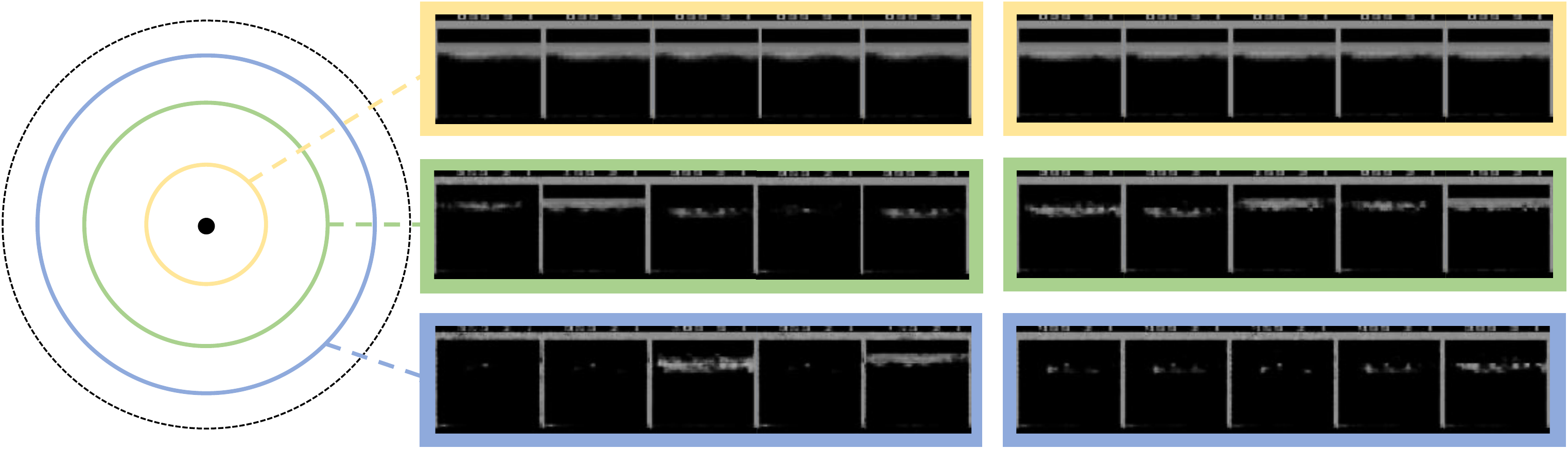}};
            \node[anchor=south west] at (0.6, 0) {\textsf{\tiny The Poincar\'e disk model}};
            \node[anchor=south west] at (5.35, 0) {\textsf{\tiny The Diagonal HWN}};
            \node[anchor=south west] at (11.1, 0) {\textsf{\tiny RoWN}};
        \end{tikzpicture}
    \caption{
    The generated images from a trained VAE endowed with RoWN by using Atari 2600 Breakout images.
    Every five images are generated from the randomly sampled latent vectors of dimensionality two having the Poincar\'e norm 0.1, 0.9, 0.95.
    The larger norm of the sample latent vectors, the more blocks are broken out in the generated images.
    }
    \label{fig:breakout_qualitative}
\end{figure}

The images of Breakout are collected by using a pre-trained Deep Q-network~\citep{mnih15} and divided into training set and test set with 90,000 and 10,000 images respectively.
We label each image with the score obtained from the game environment. 
So the labels are correlated to the number of broken blocks. 
To train VAE, we use a DCGAN-based architecture, which was originally used to evaluate HWN in \citet{nagano19}. The detailed architecture is provided in \Apxref{sec:breakout_setting}.
To evaluate the models, we measure the correlation between the Poincar\'e norm of the test images and the labeled scores. 
The evaluation results and the generated images from the trained models are reported in \autoref{tab:atari_results} and \autoref{fig:breakout_qualitative}, respectively.

In learning Breakout images, RoWN and the full covariance HWN outperform the diagonal HWN.
The isotropic HWN, where the covariance matrix is invariant to any rotation matrix, shows a better correlation than the others in all the settings.
However, as reported in \autoref{tab:atari_results_apx}, the test ELBO values are relatively worse than the others.
While the isotropic HWN shows a high correlation but relatively lower test ELBO, RoWN shows competitive test ELBO to the others and well aligns the hierarchical structures with respect to the norm in low latent dimensions.

\section{Related work}

\paragraph{Hyperbolic space for hierarchical representation learning.}
Earlier studies on the hierarchical representation learning have focused on modeling explicit hierarchical structures through the Bayesian non-parametrics~\citep{hrl3,hrl5,hrl6,hrl1,hrl4,hrl2} or embedding the hierarchical structure into Euclidean space~\citep{nickel2011three,grover2016node2vec,nguyen2017}.
Euclidean space is later shown to require an excessive number of dimensions to embed the original hierarchical structure without any distortion~\citep{linial94}.

Hyperbolic space has been proposed as an alternative medium to embed the hierarchical data. Theoretically, any tree-structured data can be embedded in hyperbolic space with arbitrarily low distortion~\citep{sarkar11}.
Based on the theory, learning representation of hierarchical structure in hyperbolic space has been shown great success in various datasets including WordNet hierarchy, graph-structured data, and social network data~\citep{nickel17, nickel18, chami20, sun20, zhao11, shavitt08}.
However, these studies are only limited to modeling the explicit hierarchy, which means that the dataset contains an explicit relation between data points. Furthermore, the learning frameworks are limited in the non-probabilistic setting because of the absence of well-behaved distribution in hyperbolic space.

\paragraph{Distributions in hyperbolic space.}
The probabilistic model enables measuring the uncertainty and provides a principled way of learning. 
To extend the probabilistic learning framework from Euclidean to hyperbolic space or Riemannian manifold in general, one needs to define a well-behaved distribution that has a tractable density and is easy to sample from.
Recently, a few studies have proposed probabilistic learning schemes in hyperbolic space~\citep{mathieu19, nagano19}.
\citet{mathieu19} introduces parametrizable sampling schemes for the two canonical Gaussian generalizations defined on the Poincaré disk model.
The scheme is used to train a hyperbolic VAE and show an improved generalization performance with high interpretability.
\citet{nagano19} suggests a method of integrating the Bayesian framework with hyperbolic space in the Lorentz model where the simpler closed form of geodesics is defined.
Normalizing flow~\citep{rezende15} can be also used to define the hyperbolic distribution in the probabilistic learning framework. 
\citet{bose2020} propose two normalizing flows defined on hyperbolic space, which show improvements in learning hierarchical structures in graph data.
\citet{mathieu2020} elevate the continuous normalizing flow~\citep{chen2018} defined on the Euclidean space to arbitrary Riemannian manifold, including the hyperbolic space.
Based on \cite{nagano19}, we analyze the geometric properties of the distribution lying on the Lorentz model and show the limitations of the existing method.
\section{Limitations}
\label{sec:lim}
We explore a better method of representing hierarchical data in hyperbolic space. To this end, we propose a simple yet effective alternative of hyperbolic wrapped normal distribution. However, the proposed distribution is limited only to hyperbolic space, and no generalization method for Riemannian space is studied yet.
To explore the usefulness of alternative Riemannian spaces, finding a common distribution that can work well in any Riemannian space will be necessary.

RoWN is a subset of the full covariance HWN. However, in many cases, RoWN outperforms the full covariance HWN in our experiments. In general, optimizing the covariance matrix requires learning the quadratic number of parameters with respect to the dimensionality. We conjecture the hardness of optimization leads to the poor performance of the full covariance HWN. To overcome this limitation, a search for a better optimization algorithm in hyperbolic space needs to be explored.

\section{Conclusions}
In this work, we propose a novel method of using RoWN for representing the data with a hierarchical structure.
With an in-depth analysis of the geometric properties of HWN, we demonstrate why the common choice of the diagonal covariance matrix for HWN may be inappropriate but the rotated covariance matrix. 
Our empirical results present that RoWN exhibits better representation ability, both qualitatively and quantitatively, compared to all the baselines: Euclidean normal distribution, diagonal HWN, and full covariance HWN.
We hope that our method helps better understanding the anatomy of hyperbolic space and be a promising technique for efficient representation learning.


\newpage
\medskip

\section*{Acknowledgement}

This work was partly supported by Institute of Information \& communications Technology Planning \& Evaluation (IITP) grant funded by the Korea government (MSIT) (No.2019-0-01906, Artificial Intelligence Graduate School Program (POSTECH)) and the National Research Foundation of Korea (NRF) grant
funded by the Korea government (MSIT) (NRF-2021R1C1C1011375).

\bibliographystyle{plainnat}
\bibliography{references.bib}

\newpage
\section*{Checklist}


\begin{enumerate}

\item For all authors...
\begin{enumerate}
  \item Do the main claims made in the abstract and introduction accurately reflect the paper's contributions and scope?
    \answerYes{}
  \item Did you describe the limitations of your work?
    \answerYes{See Section \ref{sec:lim}}
  \item Did you discuss any potential negative societal impacts of your work?
    \answerNA{}
  \item Have you read the ethics review guidelines and ensured that your paper conforms to them?
    \answerYes{}
\end{enumerate}

\item If you are including theoretical results...
\begin{enumerate}
  \item Did you state the full set of assumptions of all theoretical results?
    \answerYes{}
        \item Did you include complete proofs of all theoretical results?
    \answerYes{}
\end{enumerate}

\item If you ran experiments...
\begin{enumerate}
  \item Did you include the code, data, and instructions needed to reproduce the main experimental results (either in the supplemental material or as a URL)?
    \answerYes{The code is provided with the supplemental materials.}
  \item Did you specify all the training details (e.g., data splits, hyperparameters, how they were chosen)?
    \answerYes{}
        \item Did you report error bars (e.g., with respect to the random seed after running experiments multiple times)?
    \answerYes{}
        \item Did you include the total amount of compute and the type of resources used (e.g., type of GPUs, internal cluster, or cloud provider)?
    \answerNo{}
\end{enumerate}

\item If you are using existing assets (e.g., code, data, models) or curating/releasing new assets...
\begin{enumerate}
  \item If your work uses existing assets, did you cite the creators?
    \answerYes{}
  \item Did you mention the license of the assets?
    \answerNo{}
  \item Did you include any new assets either in the supplemental material or as a URL?
    \answerYes{}
  \item Did you discuss whether and how consent was obtained from people whose data you're using/curating?
    \answerNA{}
  \item Did you discuss whether the data you are using/curating contains personally identifiable information or offensive content?
    \answerNA{}
\end{enumerate}

\item If you used crowdsourcing or conducted research with human subjects...
\begin{enumerate}
  \item Did you include the full text of instructions given to participants and screenshots, if applicable?
    \answerNA{}
  \item Did you describe any potential participant risks, with links to Institutional Review Board (IRB) approvals, if applicable?
    \answerNA{}
  \item Did you include the estimated hourly wage paid to participants and the total amount spent on participant compensation?
    \answerNA{}
\end{enumerate}

\end{enumerate}


\appendix

\newpage
\section{Proof of Proposition 1}
\label{sec:proof1}

We prove the first proposition in \autoref{sec:observations} in this section.
Through the proposition, we first show that straight lines that pass through the origin in the Euclidean space are transformed into the geodesics in hyperbolic space by \autoref{eq:wn_operation}.
\paragraph{Proposition 1.}
Suppose $\ell_\rvs(t) = t \rvs  \in \R^{n}$ be a line passing through the origin, where $\rvs \in \R^n$ is a directional vector.
Then the curve $f_{\boldsymbol{\mu}}(\ell_\rvs(t))$ in the Lorentz model $\mathbb{L}^n$ becomes a geodesic.

\begin{proof}
Let $\rvu := [x_0(\rvu), x_{1:}(\rvu)] \in \mathbb{L}^n$ be given, where $x_0 : \mathbb{L}^n \rightarrow \R$ and $x_{1:} : \mathbb{L}^n \rightarrow \R^n$ denotes the projections, i.e., $x_i (\rvu) = u_i $ . Then, 
\begin{align*}
    \mathrm{PT}_{\boldsymbol{0}_{\mathcal{L}} \rightarrow \boldsymbol{\mu}}([0, t \rvs]) &= [0, t \rvs] + \frac{1}{x_0(\boldsymbol{\mu}) + 1} \langle \boldsymbol{\mu} - x_0(\boldsymbol{\mu}) \cdot \boldsymbol{0}_{\mathcal{L}}, [0, t \rvs] \rangle_{\mathcal{L}}(\boldsymbol{0}_{\mathcal{L}} + \boldsymbol{\mu}) \\
    &= [0, t \rvs] + \frac{1}{x_0(\boldsymbol{\mu}) + 1} \langle [0, x_{1:}(\boldsymbol{\mu})], [0, t \rvs] \rangle_{\mathcal{L}} [x_0(\boldsymbol{\mu}) + 1, x_{1:}(\boldsymbol{\mu})] \\
    &= [0, t \rvs] + \frac{1}{x_0(\boldsymbol{\mu}) + 1} \langle x_{1:}(\boldsymbol{\mu}), t \rvs \rangle [x_0(\boldsymbol{\mu}) + 1, x_{1:}(\boldsymbol{\mu})] \\
    &= t \left[\langle x_{1:}(\boldsymbol{\mu}), \rvs \rangle, \rvs + \langle x_{1:}(\boldsymbol{\mu}), \rvs \rangle \frac{x_{1:}(\boldsymbol{\mu})}{x_0(\boldsymbol{\mu}) + 1} \right].
\end{align*} 

Now, let $\rvv := \left[\langle x_{1:}(\boldsymbol{\mu}), \rvs \rangle, \rvs + \langle x_{1:}(\boldsymbol{\mu}), \rvs \rangle \frac{x_{1:}(\boldsymbol{\mu})}{x_0(\boldsymbol{\mu}) + 1} \right]$ and $c := \sqrt{\langle \rvv, \rvv \rangle_\mathcal{L}}$. Then,
\begin{align*}
    f_{\boldsymbol{\mu}}(l_\rvs) &= \exp_{\boldsymbol{\mu}}(\mathrm{PT}_{\boldsymbol{0}_{\mathcal{L}} \rightarrow \boldsymbol{\mu}}((0, t \rvs))) \\
    &= \exp_{\boldsymbol{\mu}}(t \rvv) \\
    &= \cosh(ct) \boldsymbol{\mu} + \sinh(ct) \frac{\rvv}{c} \label{eq:geodesic}.
\end{align*}
Recall that the geodesic of the Lorentz model is $\cosh(t) \rvx + \sinh(t) \rvy$ where $\rvx \in \mathbb{L}^n, \langle \rvy, \rvy \rangle_{\mathcal{L}}=1$ and $\rvy \in \mathcal{T}_\rvx \mathbb{L}^n$~\citep{robbin22}.
\end{proof}

The proposition indicates that every straight line that passes through the origin including the principal axes, is transformed into a geodesic in hyperbolic space.
\newpage
\section{Proof of Proposition 2}
\label{sec:proof2}
We prove the second proposition in \autoref{sec:observations} to show the geometrical properties of HWN.
Based on the first proposition, we provide our main proposition, which fully characterizes the structure of principal axes when projected to the Poincar\'e disk model:
\paragraph{Proposition 2.}
Let $\operatorname{Proj}(\rvu)$ be the projection function from Lorentz model to Poincar\'e model, i.e., $\operatorname{Proj}(\rvu) = \frac{x_{1:}(\rvu)}{x_0(\rvu) + 1},  \forall \rvu \in \mathbb{L}^n$. Let $\ell_\rvs$ be a principal axis of the normal distribution defined in $\mathbb{R}^n$ and $\boldsymbol{\mu}$ the mean of HWN in $\mathbb{L}^n$, then $\rvs$ is the tangent vector of $\operatorname{Proj}(f_{\boldsymbol{\mu}}(\ell_\rvs))$ on $\operatorname{Proj}(\boldsymbol{\mu})$.

\begin{proof}
First, we project the transformed principal axis to the Poincar\'e disk model as:
\begin{align*}
    \mathrm{Proj}\left(f_{\boldsymbol{\mu}}(l_\rvs)\right) &= \mathrm{Proj}\left(\cosh(c t) \boldsymbol{\mu} + \sinh(c t) \frac{\rvv}{c}\right) \\
    &= \mathrm{Proj}\left(\left[ \cosh(c t)x_0(\boldsymbol{\mu}) + \sinh(c t) \frac{x_0(\rvv)}{c}, \cosh(c t)x_{1:}(\boldsymbol{\mu}) + \sinh(c t) \frac{x_{1:}(\rvv)}{c} \right]\right) \\
    &= \frac{\cosh (c t) x_{1:}(\boldsymbol{\mu}) + \sinh(c t) x_{1:}(\rvv) / c}{\cosh(c t) x_0(\boldsymbol{\mu}) + \sinh(c t) x_0(\rvv) / c + 1}.
\end{align*}

Then, the derivative of the projected curve with respect to $t$ is derived as:
\begin{align*}
    \frac{\partial}{\partial t} \mathrm{Proj}\left(f_{\boldsymbol{\mu}}(l_\rvs)\right)  &=  \frac{\partial}{\partial t} \frac{\cosh (c t) x_{1:}(\boldsymbol{\mu}) + \sinh(c t) x_{1:}(\rvv) / c}{\cosh(c t) x_0(\boldsymbol{\mu}) + \sinh(c t) x_0(\rvv) / c + 1} \\
    &= \frac{\sinh(c t)x_{1:}(\boldsymbol{\mu})c + \cosh(c t)x_{1:}(\rvv)}{\cosh(c t)x_0(\boldsymbol{\mu}) + \sinh(c t)x_0(\rvv) / c + 1} \\
    &-\frac{(\sinh(c t)x_0(\boldsymbol{\mu}) c + \cosh(c t)x_0(\rvv))(\cosh (c t) x_{1:}(\boldsymbol{\mu}) + \sinh(c t) x_{1:}(\rvv) /c )}{\left(\cosh(c t)x_0(\boldsymbol{\mu}) + \sinh(c t)x_0(\rvv) / c + 1 \right)^2}.
\end{align*}

As $\operatorname{Proj}(\boldsymbol{\mu})$ is the point of the curve at $t=0$, the tangent vector of the curve on $\operatorname{Proj}(\boldsymbol{\mu})$ can be computed by substituting $t=0$:
\begin{align*}
    \frac{\partial}{\partial t} \mathrm{Proj}\left(f_{\boldsymbol{\mu}}(l_\rvs)\right) \Bigr|_{t = 0} &= \frac{x_{1:}(\rvv)}{x_0(\boldsymbol{\mu}) + 1} - \frac{x_0(\rvv)x_{1:}(\boldsymbol{\mu})}{(x_0(\boldsymbol{\mu}) + 1)^2} \\
    &= \frac{\rvs}{x_0(\boldsymbol{\mu}) + 1} + \frac{\langle x_{1:}(\boldsymbol{\mu}), \rvs \rangle x_{1:}(\boldsymbol{\mu})}{(x_0(\boldsymbol{\mu}) + 1)^2} - \frac{\langle x_{1:}(\boldsymbol{\mu}), \rvs \rangle x_{1:}(\boldsymbol{\mu})}{(x_0(\boldsymbol{\mu}) + 1)^2} \\
    &= \frac{\rvs}{x_0(\boldsymbol{\mu}) + 1} \\
    &\propto \rvs.
\end{align*}
\end{proof}

The proposition reveals that the principal axes of the HWN are locally parallel to the standard bases in the Poincar\'e disk model. To visualize the proposition, we plot the contour line and the principal axes of the two-dimensional diagonal covariance normal distribution before and after the transformation in \autoref{fig:wn_visualization}. 
We observe that the tangent lines of the transformed principal axes are parallel to the standard bases in hyperbolic space. 

\newpage
\section{Details of Rotated Hyperbolic Wrapped Normal Distribution}
\label{sec:rown_apx}

The details of RoWN, i.e. sampling and the probability density computation, are described in this section.
We start with a mean vector $\boldsymbol{\mu} \in \mathbb{L}^n$ and a diagonal covariance matrix $\Sigma$ as in the standard HWN.
Based on the mean vector, we construct RoWN by rotating the covariance matrix as follows:
\begin{enumerate}
    \item Compute the rotation matrix $\mR$ that rotates the x-axis ($[\pm1, \dots, 0] \in \R^n$) to $\boldsymbol{\mu}_{1:}$.
    \item Substitute the covariance matrix of Gaussian normal with $\mR \Sigma \mR^T$.
\end{enumerate}
Thus, the rotation matrix $\mR$, which rotates a unit vector from $\rvx$ to $\rvy$, can be computed as:
\begin{equation}
    \mR = \mI + (\rvy^T \rvx - \rvx^T \rvy) + \frac{1}{1 + \langle \rvx, \rvy \rangle} (\rvy^T \rvx - \rvx^T \rvy)^2.
\end{equation}
Algorithm \ref{alg:rown_construction} shows the entire algorithm of constructing RoWN.

Note that the construction is straightforward but still keeps the following benefits of the HWN: 1) The sampling can be done efficiently, and 2) the computation of the probability density of the samples is tractable. As the HWN provides a tractable probability density function for any kind of covariance matrix, we can easily compute the probability density of a given sample from RoWN. For the details of sampling and probability density computation, see Algorithm \ref{alg:rown_sampling} \& \ref{alg:rown_density}.

\begin{algorithm}[H]
\setstretch{1.3}
\caption{RoWN($\boldsymbol{\mu}, \Sigma$)}
 \textbf{Input} Mean $\boldsymbol{\mu} \in \mathbb{L}^n$, diagonal covariance matrix $\Sigma \in \R^{n \times n}$ \\
 \textbf{Output} Rotated covariance matrix $\hat{\Sigma}$.
\begin{algorithmic}[1]
\State $\rvx = [\pm 1, \dots, 0] \in \R^n, \rvy = \boldsymbol{\mu}_{1:} / \Vert \boldsymbol{\mu}_{1:} \Vert$ \Comment{$\pm$ is determined by the sign of $\boldsymbol{\mu}_0$}
\State $\mR = \mI + (\rvy^T \rvx - \rvx^T \rvy) + (\rvy^T \rvx - \rvx^T \rvy)^2 / (1 + \langle \rvx, \rvy \rangle)$
\State \textbf{return} $\hat{\Sigma} = \mR \Sigma \mR^T$
\end{algorithmic}
\label{alg:rown_construction}
\end{algorithm}

\begin{algorithm}[H]
\setstretch{1.3}
\caption{Sampling process with the rotated hyperbolic wrapped normal distribution}
 \textbf{Input} Mean $\boldsymbol{\mu} \in \mathbb{L}^n$, diagonal covariance matrix $\Sigma \in \R^{n \times n}$ \\
 \textbf{Output} Sample $\rvz \in \mathbb{L}^n$
\begin{algorithmic}[1]
\State Construct $\hat{\Sigma} = \textrm{RoWN}(\boldsymbol{\mu}, \Sigma)$
\State Sample $\rvv \sim \mathcal{N}(\boldsymbol{0}, \hat{\Sigma})$
\State \textbf{return} $\rvz = f_{\boldsymbol{\mu}}(\rvv)$
\end{algorithmic}
\label{alg:rown_sampling}
\end{algorithm}

\begin{algorithm}[H]
\setstretch{1.3}
\caption{Probability density computation of the rotated hyperbolic wrapped normal distribution}
 \textbf{Input} Mean $\boldsymbol{\mu} \in \mathbb{L}^n$, diagonal covariance matrix $\Sigma \in \R^{n \times n}$, sample $\rvz \in \mathbb{L}^n$ \\
 \textbf{Output} Log probability of $\rvz$.
\begin{algorithmic}[1]
\State Construct $\hat{\Sigma} = \textrm{RoWN}(\boldsymbol{\mu}, \Sigma)$
\State $\rvu = \log_{\boldsymbol{\mu}}(\rvz)$
\State $\rvv = \operatorname{PT}_{\boldsymbol{\mu} \rightarrow \boldsymbol{0}_{\mathcal{L}}}(\rvu)$ \Comment{$\rvv = f_{\boldsymbol{\mu}}^{-1}(\rvz)$}
\State \textbf{return} $(\textrm{log probability of} \, \rvv_{1:}\,  \textrm{from}\,  \mathcal{N}(\boldsymbol{0}, \hat{\Sigma})) - (n - 1) (\log \sinh\Vert \rvu \Vert_{\mathcal{L}} - \log \Vert \rvu \Vert_{\mathcal{L}})$
\end{algorithmic}
\label{alg:rown_density}
\end{algorithm}

\newpage

\section{Experimental Details}
\label{sec:experiments_apx}

In this section, we provide the details of the experiments in Section \ref{sec:experiments}.

\subsection{Baselines}
\label{sec:baselines}
For all the experiments, we compare the performance of RoWN with four baselines: the normal distribution in the Euclidean space, the isotropic HWN~\citep{nagano19}, the diagonal HWN, and the full covariance HWN.

In the process of constructing the distributions for each application, i.e. the variational distribution of VAE and the embedding distribution of probabilistic word embedding,
the distributions except the full covariance HWN have a diagonal matrix as an input, and then the softplus operation is used to make it positive.
The full covariance HWN has a 2D matrix $\Sigma \in \R^{n \times n}$ as an input and constructs a covariance matrix as $\Sigma \Sigma^T + \epsilon \mI$, to match the positive-definite property, where $\epsilon$ is set to $1\mathrm{e}{-9}$ in our experiments.
For the mean value, we concatenate zero at the first dimension and transport it to the Lorentz model using $\exp_{\boldsymbol{0}_{\mathcal{L}}}$.

For the hyperbolic VAE models, we use $\log_{\boldsymbol{0}_{\mathcal{L}}}$ to transform the input of the decoder to the Euclidean space, as suggested in \citet{mathieu19}.

\subsection{Noisy synthetic binary tree}
\paragraph{Experimental setting.}
A synthetic binary tree dataset is first used to show the performance of representing hierarchy in \citet{nagano19}, where each node in a tree corresponds to a sequence of binary values. 
\autoref{fig:syn_three_sub} shows an example of the depth three binary tree, where a parent and child only differ in one digit.
We add spherical noises to the nodes in the same level of hierarchy as described in \autoref{fig:syn_three_sub} as the noisy samples. With the data points with additional noises, we can create a dataset containing a local-level variation in the hierarchy. For the experiments, we uniformly sample the spherical noise from  $[0, \pi/4]$.

We train hyperbolic VAE on \textit{noisy synthetic binary tree} with varying depth from 4 to 7.
For each depth $d$, we use a three-layer fully connected neural network as the architecture where the number of hidden units is $2^{d + 3}$ and the latent dimension is $d$.
We use ReLU as the activation function for each layer except the last layer of the encoder and decoder.
The overall architecture is shown in \autoref{tab:nbt_encoder} and \autoref{tab:nbt_decoder}.
We use a Gaussian negative log-likelihood loss for the reconstruction loss with fixed $\sigma=0.01$, which is selected for sufficient reconstruction performance on the train set.

\begin{table}[h]
\begin{minipage}{.49\linewidth}
    \centering
    \caption{Encoder architecture for \textit{noisy synthetic binary tree}}
    \label{tab:nbt_encoder}
    \begin{tabular}{c c c}
        \toprule
        Layer & Output dim & Activation \\
        \midrule
        FC & $2^{d+3}$ & ReLU\\
        \midrule
        FC & $2^{d+3}$ & ReLU \\
        \midrule
        FC & $2d$ & None \\
        \bottomrule
    \end{tabular}
\end{minipage}
\hfill
\begin{minipage}{.49\linewidth}
    \centering
    \caption{Decoder architecture for \textit{noisy synthetic binary tree}}
    \label{tab:nbt_decoder}
    \begin{tabular}{c c c}
        \toprule
        Layer & Output dim & Activation  \\
        \midrule
        FC & $2^{d+3}$ & ReLU \\
        \midrule
        FC & $2^{d+3}$ & ReLU \\
        \midrule
        FC & $2^d - 1$ & None \\
        \bottomrule
    \end{tabular}
\end{minipage}
\end{table}

\paragraph{Results.}
We report 1) the correlation between the hamming distance and the embedding distance and 2) the correlation between the depth and the norm of the embeddings.
The first correlation is computed over all possible pairs of test points.
For the norm of the hyperbolic embeddings, we use the Poincar\'e norm, which can be calculated by projecting the Lorentz model embedding to the Poincar\'e disk model.

As \autoref{tab:nbt_results_apx} shows, while all the models show similar performance with respect to the ELBO, the full covariance HWN and RoWN show better performance than the diagonal HWN except depth six, outperforming the Euclidean model in every setting. RoWN preserves the depth information better than the other distributions in general.
We additionally visualize the variational mean obtained by training the tree of depth three in \autoref{fig:syn_viz}, where the hierarchical structure is well preserved in the hyperbolic embedding space.

\begin{table}[h]
    \centering
    \caption{Results of \textit{noisy synthetic binary tree}. The results are averaged over 10 runs.
    The hyperbolic models outperform the Euclidean model in all settings.
    Overall, RoWN preserves the hierarchical information better than the other distributions.}
    \label{tab:nbt_results_apx}
    \begin{tabular}{c l c c c c}
        \toprule
         & & \multicolumn{4}{c}{depth} \\
         \cmidrule{3-6}
         & & 4 & 5 & 6 & 7 \\
         \midrule
         \multirow{5}{*}{\shortstack[c]{Correlation\\w/ distance}} 
         & Euclidean & $0.748_{\pm .032}$ & $0.740_{\pm .013}$ & $0.741_{\pm .008}$ & $0.733_{\pm .014}$ \\
         & HWN (isotropic $\Sigma$) & $0.773_{\pm .030}$ & \cellcolor{yellow!70}$0.809_{\pm .016}$ & $0.798_{\pm .008}$ & $0.735_{\pm .022}$ \\
         & HWN (diagonal $\Sigma)$ & $0.814_{\pm .008}$ & $0.791_{\pm .023}$ & $0.817_{\pm .010}$ & $0.759_{\pm .025}$ \\
         & HWN (full $\Sigma$) & \cellcolor{yellow!70}$0.827_{\pm .015}$ & $0.798_{\pm .026}$ & $0.798_{\pm .010}$ & \cellcolor{yellow!70}$0.794_{\pm .014}$ \\
         & RoWN & $0.820_{\pm .015}$ & $0.807_{\pm .017}$ & \cellcolor{yellow!70}$0.822_{\pm .017}$ & $0.788_{\pm .016}$  \\
         \midrule
         \multirow{5}{*}{\shortstack[c]{Correlation\\w/ depth}} 
         & Euclidean & $0.762_{\pm .117}$ & $0.807_{\pm .038}$ & $0.712_{\pm .054}$ & $0.612_{\pm .049}$ \\
         & HWN (isotropic $\Sigma$) & $0.902_{\pm .033}$ & $0.867_{\pm .034}$ & $0.811_{\pm .029}$ & $0.602_{\pm .066}$ \\
         & HWN (diagonal $\Sigma$) & $0.918_{\pm .028}$ & $0.808_{\pm .076}$ & $0.862_{\pm .035}$ & $0.697_{\pm .076}$ \\
         & HWN (full $\Sigma$) & \cellcolor{yellow!70}$0.956_{\pm .015}$ & $0.878_{\pm .051}$ & $0.870_{\pm .033}$ & $0.815_{\pm .055}$ \\
         & RoWN & $0.930_{\pm .026}$ & \cellcolor{yellow!70}$0.911_{\pm .027}$ & \cellcolor{yellow!70}$0.901_{\pm .034}$ & \cellcolor{yellow!70}$0.827_{\pm .047}$ \\
         \midrule
         \multirow{5}{*}{Test ELBO} 
         & Euclidean & $22.591_{\pm .183}$ & $55.168_{\pm .092}$ & $124.374_{\pm .093}$ & $266.854_{\pm .199}$ \\
         & HWN (isotropic $\Sigma$) & $22.026_{\pm .201}$ & $54.054_{\pm .158}$ & $122.981_{\pm .145}$ & $265.316_{\pm .217}$ \\
         & HWN (diagonal $\Sigma$) & $22.480_{\pm .144}$ & $54.540_{\pm .117}$ & $123.444_{\pm .110}$ & $265.704_{\pm .154}$ \\
         & HWN (full $\Sigma$) & $22.371_{\pm .136}$ & $55.032_{\pm .141}$ & $124.125_{\pm .189}$ & $266.499_{\pm .112}$  \\
         & RoWN & $22.354_{\pm .138}$ & $54.648_{\pm .142}$ & $123.606_{\pm .066}$ & $266.146_{\pm .112}$ \\
         \bottomrule
    \end{tabular}
\end{table}

\begin{table}[h]
    \centering
    \caption{Correlation between the radian axis and the angular axis.}
    \label{tab:nbt_decomposition_apx}
    \begin{tabular}{c l c c c c}
        \toprule
         & & \multicolumn{4}{c}{depth} \\
         \cmidrule{3-6}
         & & 4 & 5 & 6 & 7 \\
         \midrule
         \multirow{4}{*}{\shortstack[c]{Correlation btw.\\ $r$ and $\theta_1$}} 
         & Euclidean & $0.144_{\pm .170}$ & $0.007_{\pm .105}$ & $0.039_{\pm .106}$ & $-0.026_{\pm .093}$\\
         & HWN (diagonal $\Sigma$) & $0.025_{\pm .150}$ & $0.015_{\pm .137}$ & $0.110_{\pm .065}$ & $-0.003_{\pm .134}$\\
         & HWN (full $\Sigma$) & $-0.053_{\pm .216}$ & $0.049_{\pm .136}$ & $-0.017_{\pm .169}$ & $0.012_{\pm .066}$\\
         & RoWN & $0.080_{\pm .163}$ & $0.030_{\pm .097}$ & $0.112_{\pm .093}$ & $0.025_{\pm .083}$\\
         \midrule
         \multirow{4}{*}{\shortstack[c]{Correlation btw.\\ $r$ and $\theta_2$}} 
         & Euclidean & $0.116_{\pm .232}$ & $-0.039_{\pm .210}$ & $0.109_{\pm .131}$ & $0.006_{\pm .095}$\\
         & HWN (diagonal $\Sigma$) & $0.066_{\pm .170}$ & $-0.021_{\pm .190}$ & $0.044_{\pm .122}$ & $-0.024_{\pm .115}$\\
         & HWN (full $\Sigma$) & $0.297_{\pm .116}$ & $0.113_{\pm .109}$ & $0.013_{\pm .122}$ & $0.061_{\pm .103}$\\
         & RoWN & $0.013_{\pm .178}$ & $0.025_{\pm .173}$ & $-0.004_{\pm .115}$ & $0.064_{\pm .082}$\\
         \midrule
         \multirow{4}{*}{\shortstack[c]{Correlation btw.\\ $r$ and $\theta_3$}} 
         & Euclidean & $0.067_{\pm .220}$ & $-0.110_{\pm .220}$ & $-0.016_{\pm .159}$ & $0.011_{\pm .098}$\\
         & HWN (diagonal $\Sigma$) & $0.123_{\pm .252}$ & $-0.139_{\pm .123}$ & $-0.019_{\pm .133}$ & $-0.095_{\pm .101}$\\
         & HWN (full $\Sigma$) & $-0.053_{\pm .144}$ & $-0.088_{\pm .107}$ & $0.106_{\pm .120}$ & $0.024_{\pm .079}$\\
         & RoWN & $0.127_{\pm .253}$ & $-0.120_{\pm .169}$ & $-0.012_{\pm .120}$ & $-0.012_{\pm .083}$\\
         \midrule
         \multirow{4}{*}{\shortstack[c]{Correlation btw.\\ $r$ and $\theta_4$}} 
         & Euclidean & - & $-0.015_{\pm .073}$ & $0.013_{\pm .081}$ & $0.026_{\pm .065}$\\
         & HWN (diagonal $\Sigma$) & - & $-0.047_{\pm .117}$ & $-0.035_{\pm .147}$ & $0.080_{\pm .096}$\\
         & HWN (full $\Sigma$) & - & $0.079_{\pm .150}$ & $0.042_{\pm .108}$ & $0.086_{\pm .102}$\\
         & RoWN & - & $-0.031_{\pm .116}$ & $-0.070_{\pm .112}$ & $0.086_{\pm .115}$\\
         \midrule
         \multirow{4}{*}{\shortstack[c]{Correlation btw.\\ $r$ and $\theta_5$}} 
         & Euclidean & - & - & $0.082_{\pm .075}$ & $-0.029_{\pm .109}$\\
         & HWN (diagonal $\Sigma$) & - & - & $0.022_{\pm .101}$ & $-0.041_{\pm .097}$\\
         & HWN (full $\Sigma$) & - & - & $-0.058_{\pm .111}$ & $0.076_{\pm .064}$\\
         & RoWN & - & - & $-0.016_{\pm .111}$ & $-0.030_{\pm .112}$\\
         \midrule
         \multirow{4}{*}{\shortstack[c]{Correlation btw.\\ $r$ and $\theta_6$}} 
         & Euclidean & - & - & - & $-0.018_{\pm .073}$\\
         & HWN (diagonal $\Sigma$) & - & - & - & $-0.030_{\pm .082}$\\
         & HWN (full $\Sigma$) & - & - & - & $0.035_{\pm .103}$\\
         & RoWN & - & - & - & $0.006_{\pm .067}$\\
         \bottomrule
    \end{tabular}
\end{table}

\paragraph{Decomposition of the radial and angular dependency.}
To show how well the angular and radial representations are decomposed, we compute the Pearson correlation between the radial axis and the angular axis, which has been shown to be an effective measure of variable dependency in disentangled representation learning \citep{jo2019,horan2021}.
\autoref{tab:nbt_decomposition_apx} shows that the absolute value of the correlation is lower or similar to 0.1 in all the models, including RoWN.

\begin{figure}[b]
    \centering
    \includegraphics[width=.3\linewidth]{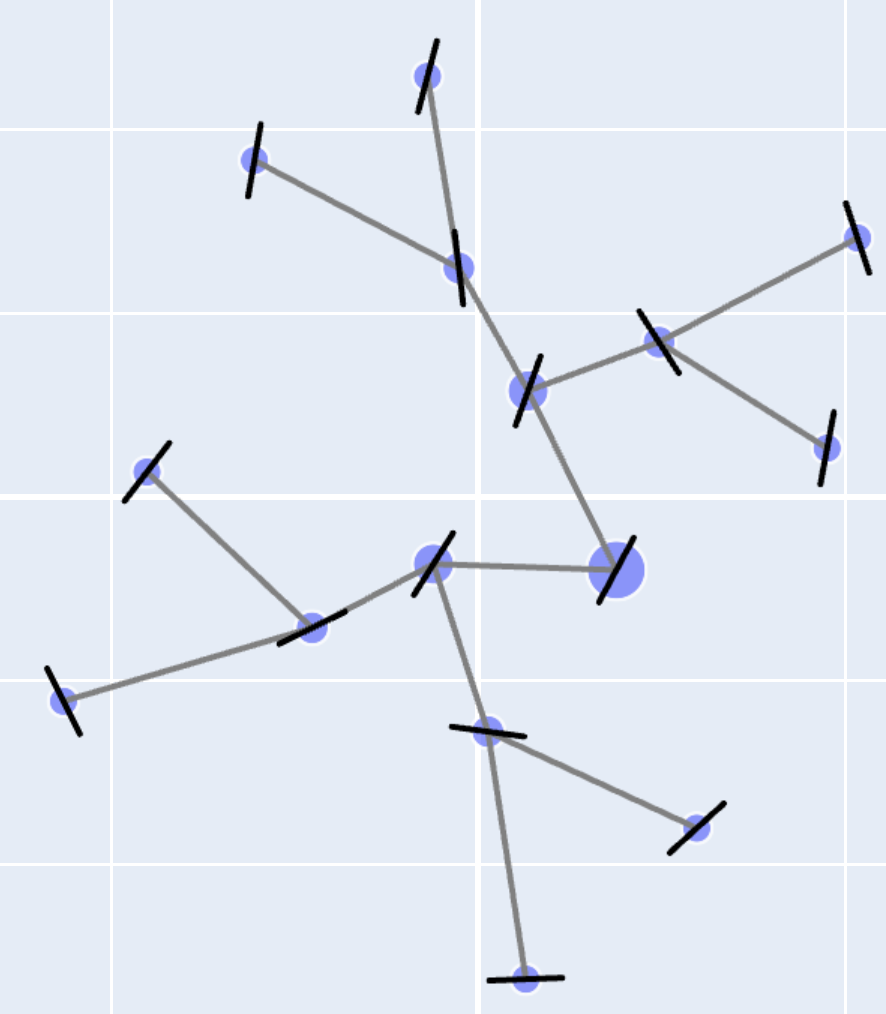}
    \caption{Visualization of the representations from learnable rotation model. The representations are from the Learnable Rotation 1 model learned on the depth 4 noisy synthetic binary tree with latent dimension 2. The size of the circles denotes the depth, where the biggest circle denotes the root node. Most of the rotation directions are pointing or orthogonal (black line segments on each node) to the direction of its parent node.}
    \label{fig:learnable_rotation}
\end{figure}

\paragraph{Learnable rotation.}
We add experiments for the models that learn the rotation direction, which is originally fixed to the direction of $\boldsymbol{\mu}$ (in Algorithm 1, the $\rvy$ vector). 
Given data, the encoder gives the rotation direction. We test the models on the noisy synthetic binary tree setting in our paper. 
As shown \autoref{tab:nbt_results_learnable_x_apx}, as the depth becomes deeper, the learnable rotation models usually underperform RoWN.
\autoref{fig:learnable_rotation} shows behaviors of learned representation by an alternative Learnable Rotation 1. Most of the rotation directions are pointing or orthogonal (black line segments on each node) to the direction of its parent node. This implies that the alternatives of RoWN can learn representations of nodes to align not to the root node but to the parent node. We note that these alternatives work well with shallow depths but not with great depths.

\begin{table}[h]
    \centering
    \caption{Results of learnable rotation models. Learnable Rotation 1 model outputs the rotation direction parallel to the mean and variance, while Learnable Rotation 2 model outputs the rotation direction by feeding the mean to a fully connected layer.}
    \label{tab:nbt_results_learnable_x_apx}
    \begin{tabular}{c l c c c c}
        \toprule
         & & \multicolumn{4}{c}{depth} \\
         \cmidrule{3-6}
         & & 4 & 5 & 6 & 7 \\
         \midrule
         \multirow{3}{*}{\shortstack[c]{Correlation\\w/ distance}} 
         & RoWN & $0.820_{\pm .015}$ & $0.807_{\pm .017}$ & $0.822_{\pm .017}$ & $0.788_{\pm .016}$ \\
         & Learnable Rotation 1 & $0.820_{\pm .013}$ & $0.804_{\pm .016}$ & $0.808_{\pm .013}$ & $0.770_{\pm .028}$ \\
         & Learnable Rotation 2 & $0.817_{\pm .018}$ & $0.811_{\pm .018}$ & $0.801_{\pm .008}$ & $0.772_{\pm .021}$ \\
         \midrule
         \multirow{3}{*}{\shortstack[c]{Correlation\\w/ depth}} 
         & RoWN & $0.930_{\pm .026}$ & $0.911_{\pm .027}$ & $0.901_{\pm .034}$ & $0.827_{\pm .047}$\\
         & Learnable Rotation 1 & $0.952_{\pm .018}$ & $0.903_{\pm .037}$ & $0.845_{\pm .057}$ & $0.711_{\pm .072}$\\
         & Learnable Rotation 2 & $0.948_{\pm .015}$ & $0.907_{\pm .019}$ & $0.844_{\pm .013}$ & $0.724_{\pm .058}$\\
         \midrule
         \multirow{3}{*}{Test ELBO} 
         & RoWN & $22.354_{\pm .138}$ & $54.648_{\pm .142}$ & $123.606_{\pm .066}$ & $266.146_{\pm .112}$\\
         & Learnable Rotation 1 & $22.342_{\pm .097}$ & $54.741_{\pm .111}$ & $123.619_{\pm .180}$ & $266.076_{\pm .159}$\\
         & Learnable Rotation 2 & $22.286_{\pm .120}$ & $54.529_{\pm .067}$ & $123.382_{\pm .161}$ & $265.962_{\pm .217}$\\
         \bottomrule
    \end{tabular}
\end{table}

\subsection{WordNet}
\paragraph{Experimental setting.}
We train a probabilistic word embedding model with WordNet dataset~\citep{fellbaum98}.
We initialize the mean and variance parameters with $\mathcal{N}(\boldsymbol{0}, 0.01)$.
For the full covariance model, we use a learning rate $0.01$.
For the other models, we set the learning rate $0.015$ for the first 100 epochs and then set the learning rate to $0.6$ for the remaining steps as done in \citep{nickel17, nagano19}.
We evaluate the learned representations by computing the average rank of all the hypernymies. 
The rank of a given pair of words $s$ and $t$ is computed among the distances between all possible pairs of the words $s$ and $t'$ without hypernymy.

\paragraph{Results.}

We evaluate the learned representations by computing the average rank of all the hypernymies.
The rank of a given pair of words $s$ and $t$ is computed among the distances between all possible pairs of the words $s$ and $t'$ without hypernymy.
\autoref{tab:wordnet_results} shows the empirical performances of representing the word data. We report the performance with the mean rank (MR) and the mean average precision (mAP).
RoWN preserves the hierarchical structure better than the other distributions, while the full covariance HWN fails due to unstable optimization.

\begin{table}[b!]
    \centering
    \caption{Results of \textit{WordNet}. 
    The results are an average of 5 runs.
    Based on the rank of hypernymy pairs among non-hypernymy pairs, we report the mean rank (MR) and mean average precision (mAP) for evaluation.}
    \label{tab:wordnet_results_apx}
    \vspace{1em}
    \begin{tabular}{c l c c c}
        \toprule
         & & \multicolumn{3}{c}{latent dimension} \\
         \cmidrule{3-5}
         &  & 5 & 10 & 20 \\
         \midrule
         \multirow{5}{*}{MR} 
         & Euclidean & \cellcolor{yellow!70}$13.968_{\pm 0.504}$ & $3.862_{\pm 0.281}$ & $1.955_{\pm 0.157}$ \\
         & HWN (isotropic $\Sigma$) & $14.568_{\pm 2.203}$ & $4.470_{\pm 0.669}$ & $3.125_{\pm 0.455}$\\
         & HWN (diagonal $\Sigma$) & $16.590_{\pm 1.146}$ & $3.891_{\pm 0.447}$ & $2.062_{\pm 0.088}$\\
         & HWN (full $\Sigma$) & $557.309_{\pm 18.006}$ & $466.513_{\pm 75.142}$ & $599.140_{\pm 18.916}$ \\
         & RoWN & $16.271_{\pm 2.985}$ & \cellcolor{yellow!70}$2.888_{\pm 0.162}$ & \cellcolor{yellow!70}$1.783_{\pm 0.090}$\\
         \midrule
         \multirow{5}{*}{mAP} 
         & Euclidean & $0.565_{\pm 0.014}$ & $0.801_{\pm 0.020}$ & $0.902_{\pm 0.008}$\\
         & HWN (isotropic $\Sigma$) & \cellcolor{yellow!70}$0.617_{\pm 0.012}$ & $0.820_{\pm 0.013}$ & $0.847_{\pm 0.017}$\\
         & HWN (diagonal $\Sigma$)& $0.565_{\pm 0.020}$ & $0.805_{\pm 0.015}$ & $0.905_{\pm 0.007}$\\
         & HWN (full $\Sigma$) & $0.032_{\pm 0.003}$ & $0.063_{\pm 0.005}$ & $0.079_{\pm 0.021}$ \\
         & RoWN & $0.593_{\pm 0.024}$ & \cellcolor{yellow!70}$0.844_{\pm 0.009}$ & \cellcolor{yellow!70}$0.921_{\pm 0.005}$\\
         \bottomrule
    \end{tabular}
\end{table}

\paragraph{Optimization issue in training full covariance HWN.}

\begin{figure}
    \centering
    \begin{subfigure}[t]{.23\linewidth}
    \centering
    \begin{tikzpicture}
        \begin{axis}[
            legend style={nodes={scale=0.45, transform shape}},
            legend pos=north east,
            xmode=log,
            xlabel={learning rate},
            ylabel={MR},
            width=1.3\linewidth,
            height=1.3\linewidth,
            xlabel style={yshift=0.15cm},
            ylabel style={yshift=-0.2cm},
        ]
            \addplot[red, very thick, mark=o, mark size=2pt, mark options={solid}] table[col sep=comma, x=lr, y=c1]{results/opt_rank_full.csv};
            \addplot[green, very thick, mark=triangle, mark size=2pt, mark options={solid}] table[col sep=comma, x=lr, y=c40]{results/opt_rank_full.csv};
            \addplot[blue, very thick, mark=diamond, mark size=2pt, mark options={solid}] table[col sep=comma, x=lr, y=c100]{results/opt_rank_full.csv};
            \addplot[dashed] coordinates {(0.0001, 16.271) (0.6, 16.271)};
            \legend{$c=1$,$c=40$,$c=100$,RoWN}
        \end{axis}
    \end{tikzpicture}
    \caption{MR of full initialization}
    \end{subfigure}
    \hspace{2mm}
    \begin{subfigure}[t]{.23\linewidth}
    \centering
    \begin{tikzpicture}
        \begin{axis}[
            xmode=log,
            xlabel={learning rate},
            ylabel={mAP},
            width=1.3\linewidth,
            height=1.3\linewidth,
            xlabel style={yshift=0.15cm},
            ylabel style={yshift=-0.15cm},
        ]
            \addplot[red, very thick, mark=o, mark size=2pt, mark options={solid}] table[col sep=comma, x=lr, y=c1]{results/opt_map_full.csv};
            \addplot[green, very thick, mark=triangle, mark size=2pt, mark options={solid}] table[col sep=comma, x=lr, y=c40]{results/opt_map_full.csv};
            \addplot[blue, very thick, mark=diamond, mark size=2pt, mark options={solid}] table[col sep=comma, x=lr, y=c100]{results/opt_map_full.csv};
            \addplot[dashed] coordinates {(0.0001, 0.593) (0.6, 0.593)};
        \end{axis}
    \end{tikzpicture}
    \caption{mAP of full initialization}
    \end{subfigure}
    \hspace{2mm}
    \begin{subfigure}[t]{.23\linewidth}
    \centering
    \begin{tikzpicture}
        \begin{axis}[
            xmode=log,
            xlabel={learning rate},
            ylabel={MR},
            width=1.3\linewidth,
            height=1.3\linewidth,
            xlabel style={yshift=0.15cm},
            ylabel style={yshift=-0.2cm},
        ]
            \addplot[red, very thick, mark=o, mark size=2pt, mark options={solid}] table[col sep=comma, x=lr, y=c1]{results/opt_rank_diagonal.csv};
            \addplot[green, very thick, mark=triangle, mark size=2pt, mark options={solid}] table[col sep=comma, x=lr, y=c40]{results/opt_rank_diagonal.csv};
            \addplot[blue, very thick, mark=diamond, mark size=2pt, mark options={solid}] table[col sep=comma, x=lr, y=c100]{results/opt_rank_diagonal.csv};
            \addplot[dashed] coordinates {(0.0001, 16.271) (0.6, 16.271)};
        \end{axis}
    \end{tikzpicture}
    \caption{MR of diagonal initialization}
    \end{subfigure}
    \hspace{2mm}
    \begin{subfigure}[t]{.23\linewidth}
    \centering
    \begin{tikzpicture}
        \begin{axis}[
            xmode=log,
            xlabel={learning rate},
            ylabel={mAP},
            width=1.3\linewidth,
            height=1.3\linewidth,
            xlabel style={yshift=0.15cm},
            ylabel style={yshift=-0.2cm},
        ]
            \addplot[red, very thick, mark=o, mark size=2pt, mark options={solid}] table[col sep=comma, x=lr, y=c1]{results/opt_map_diagonal.csv};
            \addplot[green, very thick, mark=triangle, mark size=2pt, mark options={solid}] table[col sep=comma, x=lr, y=c40]{results/opt_map_diagonal.csv};
            \addplot[blue, very thick, mark=diamond, mark size=2pt, mark options={solid}] table[col sep=comma, x=lr, y=c100]{results/opt_map_diagonal.csv};
            \addplot[dashed] coordinates {(0.0001, 0.593) (0.6, 0.593)};
        \end{axis}
    \end{tikzpicture}
    \caption{mAP of diagonal initialization}
    \end{subfigure}
    \hfill
    \caption{Varying hyperparameters of the full covariance HWN on WordNet. We run several hyper-parameters combination of full covariance HWN on WordNet. We fix the latent dimension to 5 and burn-in epochs to 100. We vary the learning rate from $1\mathrm{e}{-4}$ to 0.6 and the factor $c$, which is used to reduce the learning rate in the burn-in steps, from 1 to 100. We run 10,000 epochs. (a,b) When we initialize the entire covariance matrix with $\mathcal{N}(0, 0.01)$, the full covariance HWN show poor performance with any hyper-parameter combinations. (c,d) Initializing only the diagonal entries with $\mathcal{N}(0, 0.01)$ and let the remaining part to zero improves the performance of the full covariance HWN but it still performs worse than RoWN.}
    \label{fig:optimization}
\end{figure}
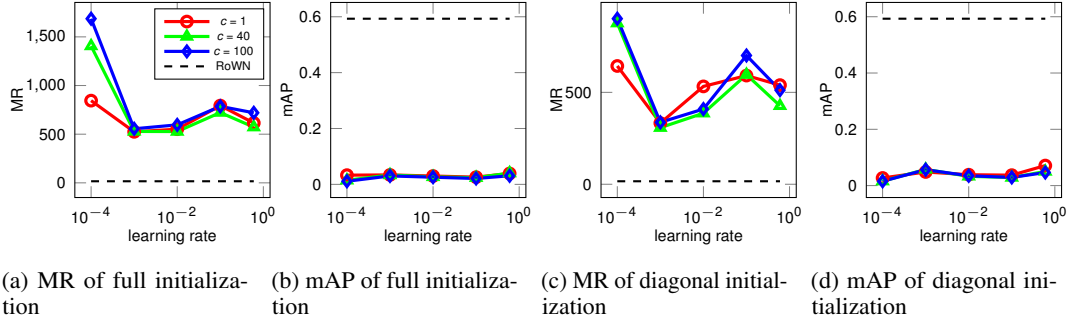

In the results, we find that the full covariance HWN shows poor performance compared to the other models.
We run extensive experiments to show that the full covariance HWN is difficult to optimize especially in WordNet.
\autoref{fig:optimization} shows the performance of the full covariance HWN with varying hyperparameters, i.e. learning rate, the burn-in factor, and the initialization method, which seems to be poor whatever we choose. 
We conducted an additional analysis on what causes the optimization instability and found that the number of samples used to approximate the KL divergence is critical to full covariance HWN, especially in the Wordnet dataset. In VAE, we can observe more stable results. We speculate that the stability improved since the relatively simple prior (standard normal distribution) is employed with the full covariance variational distribution.
\autoref{tab:wordnet_train_samples} shows that as the number of training samples increases, the performance increases, but the result is still poor, and using more training samples leads to an additional computation time.

\begin{table}[h]
    \caption{Varying samples of the full covariance HWN on WordNet.}
    \centering
    \begin{tabular}{c c c c}
        \toprule
        \# of samples & 1 & 50 & 100 \\
        \midrule
        MR & $326.185_{\pm 7.610}$ & $73.572_{\pm 9.926}$ & $69.964_{\pm 10.457}$ \\
        mAP & $0.056_{\pm 0.004}$ & $0.241_{\pm 0.021}$ & $0.241_{\pm 0.019}$ \\
        Runtime (s/epoch) & 6.288 & 6.748 & 8.163 \\
        \bottomrule
    \end{tabular}
    \label{tab:wordnet_train_samples}
\end{table}

\subsection{Atari 2600 Breakout}
\paragraph{Experimental setting.}
\label{sec:breakout_setting}
To learn the implicit hierarchy that can be observed from the trajectories of Breakout, we train the VAE models with the Atari 2600 Breakout images.
The images of Breakout are collected by using a pre-trained Deep Q-network~\citep{mnih15} and divided into a training set and test set with 90,000 and 10,000 images respectively.
We label each image with the score obtained from the game environment.  
So the labels are correlated to the number of broken blocks. 
To train VAE, we use a DCGAN-based architecture, which was originally used to evaluate HWN in \citet{nagano19}. 
The detailed architecture is provided in \autoref{tab:breakout_encoder} and \autoref{tab:breakout_decoder}.
We use binary cross-entropy loss for the reconstruction loss.

\begin{table}[h]
\begin{minipage}{.49\linewidth}
    \centering
    \caption{Encoder architecture for Breakout}
    \label{tab:breakout_encoder}
    \begin{tabular}{c c c}
        \toprule
        Layer & Output dim & Activation \\
        \midrule
        Conv2d & $80\times80\times16$ & ReLU\\
        \midrule
        Conv2d & $40\times40\times32$ & ReLU\\
        \midrule
        Conv2d & $40\times40\times32$ & ReLU\\
        \midrule
        Conv2d & $20\times20\times64$ & ReLU\\
        \midrule
        Conv2d & $20\times20\times64$ & ReLU\\
        \midrule
        Conv2d & $10\times10\times64$ & ReLU\\
        \midrule
        FC & $2\times \textrm{latent dimension}$ & None \\
        \bottomrule
    \end{tabular}
\end{minipage}
\hfill
\begin{minipage}{.49\linewidth}
    \centering
    \caption{Decoder architecture for Breakout}
    \label{tab:breakout_decoder}
    \begin{tabular}{c c c}
        \toprule
        Layer & Output dim & Activation  \\
        \midrule
        FC & $10\times10\times64$ & ReLU \\
        \midrule
        ConvTranspose2d & $20\times20\times32$ & ReLU \\
        \midrule
        Conv2d & $20\times20\times32$ & ReLU \\
        \midrule
        ConvTranspose2d & $40\times40\times16$ & ReLU \\
        \midrule
        Conv2d & $40\times40\times16$ & ReLU \\
        \midrule
        ConvTranspose2d & $80\times80\times1$ & Sigmoid \\
        \bottomrule
    \end{tabular}
\end{minipage}
\end{table}

\paragraph{Results.}
\begin{table}[h]
    \centering
    \caption{Results of \textit{Atari 2600 Breakout}. 
    The results are averaged over 10 runs.
    We measure the correlation between the score of an image and the Poincar\'e norm of the variational mean.}
    \label{tab:atari_results_apx}
    \begin{tabular}{c l c c c}
        \toprule
         & & \multicolumn{3}{c}{latent dimension} \\
         \cmidrule{3-5}
         & & 10 & 15 & 20 \\
         \midrule
         \multirow{5}{*}{\shortstack[c]{Correlation btw.\\score and norm}} 
         & Euclidean & $0.379_{\pm .007}$ & $0.436_{\pm .029}$ & $0.479_{\pm .020}$\\
         & HWN (isotropic $\Sigma$) & \cellcolor{yellow!70}$0.513_{\pm .012}$ & \cellcolor{yellow!70}$0.598_{\pm .021}$ & \cellcolor{yellow!70}$0.607_{\pm .015}$ \\
         & HWN (diagonal $\Sigma$) & $0.478_{\pm .011}$ & $0.513_{\pm .006}$ & $0.513_{\pm .008}$ \\
         & HWN (full $\Sigma$) & $0.483_{\pm .011}$ & $0.520_{\pm .009}$ & $0.563_{\pm .010}$ \\
         & RoWN & $0.497_{\pm .014}$ & $0.556_{\pm .014}$ & $0.561_{\pm .029}$ \\
         \midrule
         \multirow{5}{*}{Test ELBO} & Euclidean & $-1269.044_{\pm .241}$ & $-1269.624_{\pm .258}$ & $-1269.682_{\pm .178}$ \\
         & HWN (isotropic $\Sigma$) & $-1271.018_{\pm .440}$ & $-1272.139_{\pm .170}$ & $-1272.914_{\pm .118}$ \\
         & HWN (diagonal $\Sigma$) & $-1269.816_{\pm .272}$ & $-1270.725_{\pm .260}$ & $-1271.087_{\pm .234}$ \\
         & HWN (full $\Sigma$) & $-1269.021_{\pm .320}$ & $-1269.569_{\pm .206}$ & $-1269.882_{\pm .438}$ \\
         & RoWN & $-1269.531_{\pm .212}$ & $-1270.203_{\pm .211}$ & $-1270.967_{\pm .183}$ \\
         \bottomrule
    \end{tabular}
\end{table}


To evaluate the models, we measure the correlation between the norm of the test images and the labeled scores. 
For the norm of the hyperbolic embeddings, we use the Poincar\'e norm, which can be calculated by projecting the Lorentz model embedding to the Poincar\'e disk model.
The results are reported in \autoref{tab:atari_results_apx}.
While all the models show similar representation power with respect to ELBO, RoWN and the full covariance HWN outperform the diagonal HWN. Especially, RoWN aligns the hierarchical structures better with respect to the norm in low latent dimensions.

\citet{nagano19} report a higher score of correlation in latent dimension 20, but we find some issues with the result.
First, \citet{nagano19} compute the correlation between the labeled scores and the norm in the tangent space ($\rvv$ vector from Algorithm \ref{alg:rown_density}), not the Poincar\'e norm.
The projection function in Proposition \ref{proposition:tangent} depends on the first element of the input vector.
Thus the Poincar\'e norm is not proportional to the $\rvv$ norm, and computing the correlation with the $\rvv$ norm will show different behavior compared to the correlation with the Poincar\'e norm.
Second, the reproduction results obtained from the code by the official repository\footnote{\url{https://github.com/pfnet-research/hyperbolic_wrapped_distribution}} are far from the reported correlation.
Our reproduction results of the correlation between the labeled scores and the $\rvv$ norm show 0.616, and between the Poincar\'e norm show 0.501 averaged over four runs.

\begin{figure}[h]
    \centering
    \begin{subfigure}[t]{.49\linewidth}
        \centering
        \includegraphics[width=\linewidth]{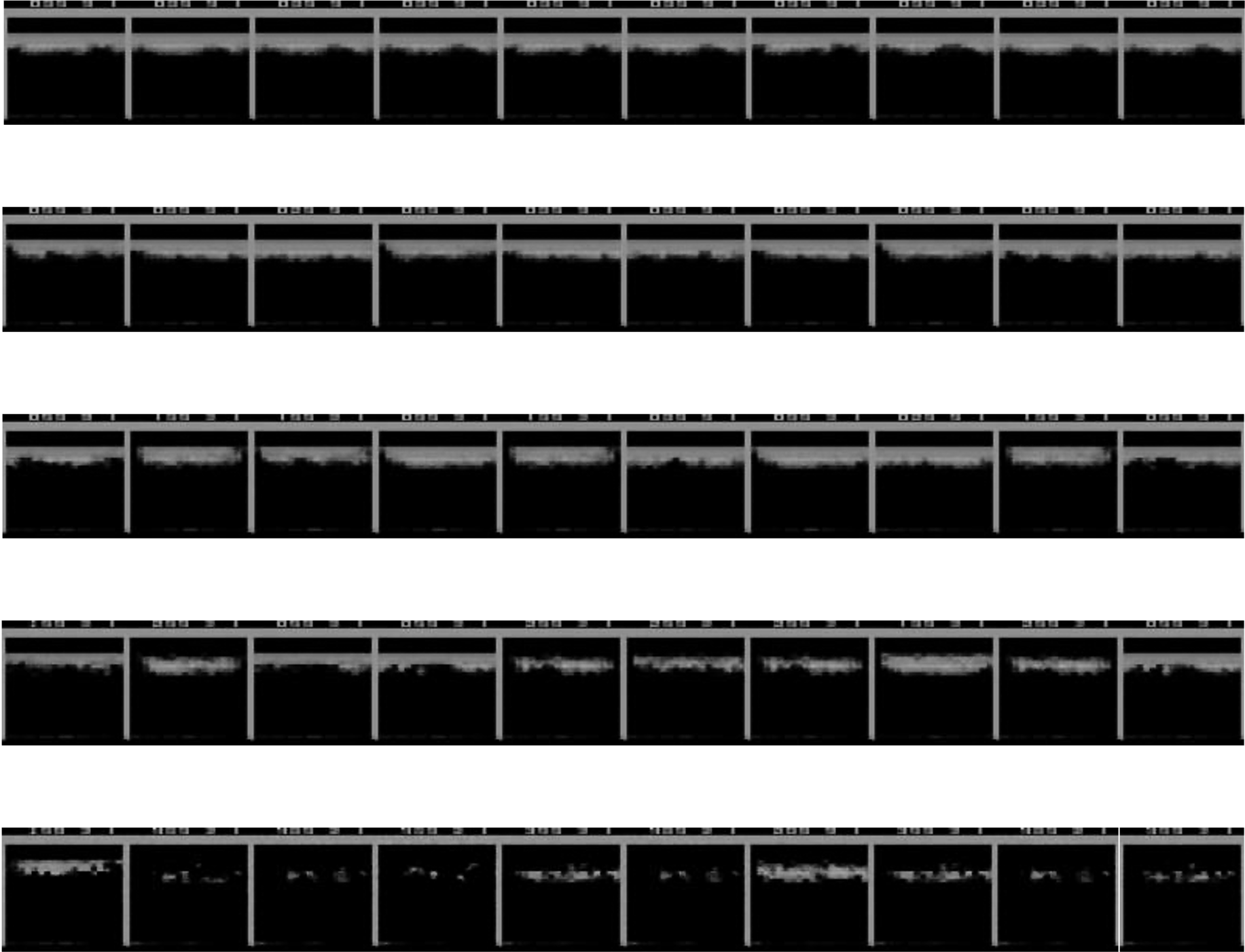}
        \caption{Diagonal HWN}
    \end{subfigure}
    \hfill
    \begin{subfigure}[t]{.49\linewidth}
        \centering
        \includegraphics[width=\linewidth]{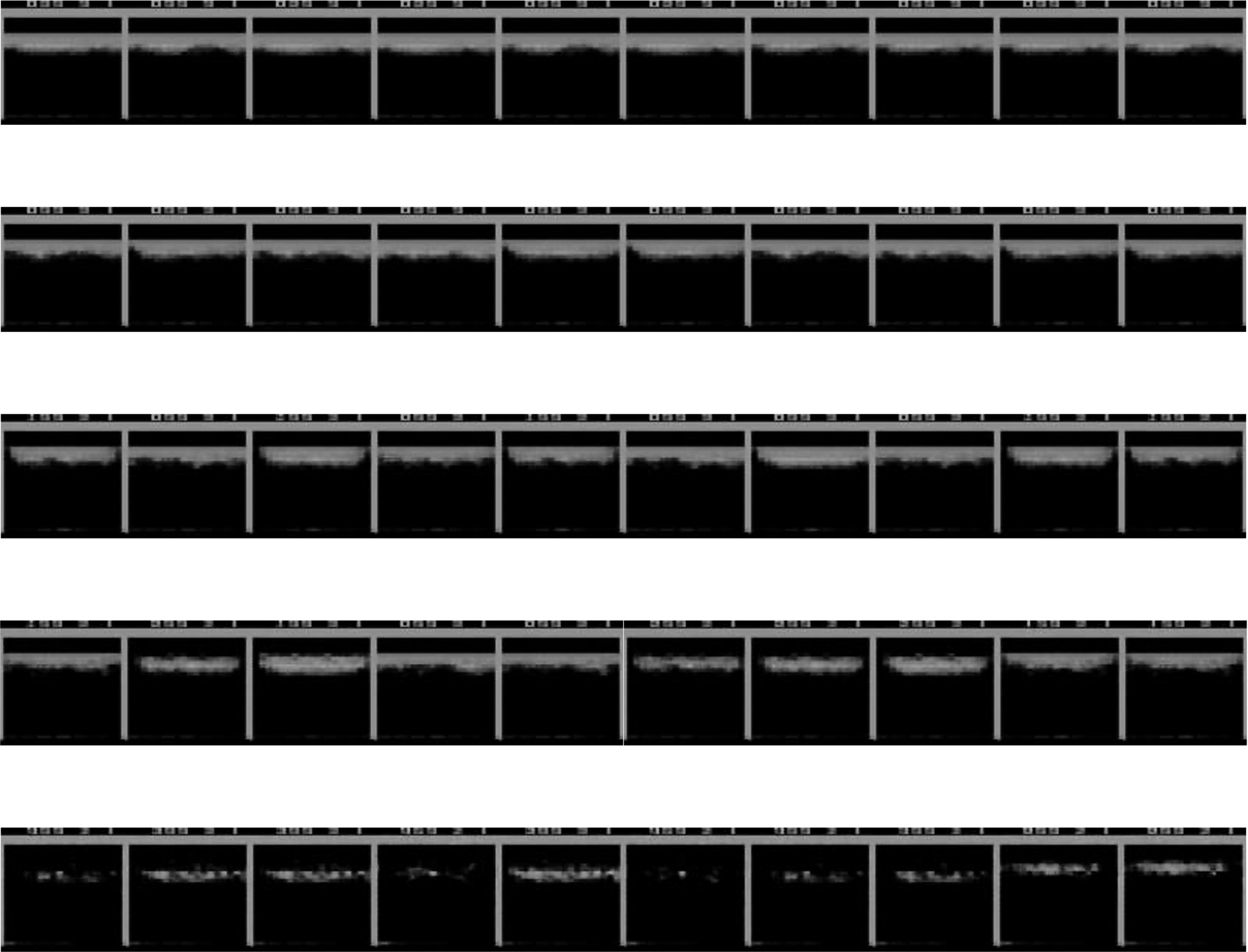}
        \caption{RoWN}
    \end{subfigure}
    \caption{Generation results of the VAE models trained on Breakout images. The models are trained on Breakout images with two dimensional latent space. We generate the images from randomly sampled latent vectors having Poincar\'e norm as 0.1, 0.3, 0.5, 0.7, 0.9.}
    \label{fig:breakout_qualitative_apx}
\end{figure}

\paragraph{Qualitative results.}
\autoref{fig:breakout_qualitative_apx} shows more examples of generated images from VAE models trained on Breakout images with two dimensional latent space.

\newpage
\section{Discussion on Root Placement}
\label{sec:root_placement}
Note that due to the isometry of the geometry, the root node can be placed anywhere in hyperbolic space. 
In other words, infinitely many sets of embeddings preserve the same pairwise distances between the nodes.
This reveals that finding the appropriate isometry, where the root node is placed near the origin, is important for using RoWN as the distribution.
In this section, we discuss the techniques we used to place the root node near the origin of each application.

\subsection{Probabilistic word embedding model}
In our experiments, to place the root node near the origin, we have initialized embeddings from $\mathcal{N}(0, 0.01 I)$, which are then moved to the Lorentz model using the exponential map, with learning rate warm-up~\cite{nagano19}. 
The results with different initialization method are shown in \autoref{tab:root_node_results}.

\subsection{Hyperbolic VAE}
When RoWN is used as a variational distribution in the hyperbolic VAE, the application of RoWN is different from the probabilistic word embedding model since the KL divergence encourages all variational means to be close to the prior mean.
Suppose we only focus on nodes at a certain depth in a tree. In that case, it can be easily identified that it would be beneficial to have all nodes at the same level of the norm to minimize the geometric mean between the prior and posterior means. Here, we assume that each pair of nodes requires to have a certain amount of distance to minimize the reconstruction error. 
As shown in \autoref{fig:hwn_plots}, the original HWN is difficult to have the nodes at the same depth with similar norms since the local variation cannot be modeled through the radial direction.
Eventually, the root node slightly deviates from the prior mean. 
With RoWN, as shown in \autoref{fig:rown_plots}, all the nodes at the same depth can be placed at a similar norm while preserving their local variations. 
If this is indeed the case, the root node is likely to be placed near the prior mean since the nodes with different depths will be placed at different levels of norms in the space.

\end{document}